\documentclass[lettersize,journal]{IEEEtran}
\usepackage{amsmath,amsfonts}
\usepackage{algorithmic}
\usepackage{algorithm}
\usepackage{array}
\usepackage[caption=false,font=normalsize,labelfont=sf,textfont=sf]{subfig}
\usepackage{textcomp}
\usepackage{stfloats}
\usepackage{url}
\usepackage{verbatim}
\usepackage{graphicx}
\usepackage{multirow}%
\usepackage{cite}
\usepackage{color}
\begin{document}

\title{Cognitive Manipulation: Semi-supervised Visual Representation and Classroom-to-real Reinforcement Learning for Assembly in Semi-structured Environments}
%fine-tuning of object detection and residual reinforcement learning of contact-rich manipulation
%knowledge-informed visual perception for robot learning in fixtureless assembly task
%Cognitive Manipulation: Skill Graph-based Semi-supervised Object Detection and Classroom-to-real residual reinforcement learning for Assembly in Semi-structured Environments

\author{Chuang Wang, Lie Yang, Ze Lin, Yizhi Liao, Gang Chen, and Longhan Xie, ~\IEEEmembership{Member, IEEE}
        % <-this % stops a space
\thanks{This document is the results of the National Key Research and Development Program of China (Grant No. 2021YFB3301400), the research project funded by the National Natural Science Foundation of China (Grant No. 52075177). (Corresponding author: Gang Chen and Longhan Xie.)}% <-this % stops a space
\thanks{Chuang Wang is with the South China University of Technology, Guangzhou, CN (e-mail: wichuangwang@mail.scut.edu.cn).}
\thanks{Lie Yang is with the School of Mechanical and Aerospace Engineering, Nanyang Technological University, Singapore. (e-mail: lie.yang@ntu.edu.sg).}
\thanks{Gang Chen is with the South China University of Technology, Guangzhou, CN (e-mail: gangchen@scut.edu.cn).}
\thanks{Longhan Xie is with the South China University of Technology, Guangzhou, CN (e-mail: melhxie@scut.edu.cn).}}

% The paper headers
\markboth{Wang \MakeLowercase{\textit{et al.}}: Cognitive Manipulation for robotic assembly in semi-structured environments}%Journal of \LaTeX\ Class Files,~Vol.~14, No.~8, August~2021
{Wang \MakeLowercase{\textit{et al.}}: Cognitive Manipulation for robotic assembly in semi-structured environments}

%\IEEEpubid{0000--0000/00\$00.00~\copyright~2021 IEEE}
% Remember, if you use this you must call \IEEEpubidadjcol in the second
% column for its text to clear the IEEEpubid mark.

\maketitle

\begin{abstract}
Assembling a slave object into a fixture-free master object represents a critical challenge in flexible manufacturing. Existing deep reinforcement learning-based methods, while benefiting from visual or operational priors, often struggle with small-batch precise assembly tasks due to their reliance on insufficient priors and high-costed model development. To address these limitations, this paper introduces a cognitive manipulation and learning approach that utilizes skill graphs to integrate learning-based object detection with fine manipulation models into a cohesive modular policy. This approach enables the detection of the master object from both global and local perspectives to accommodate positional uncertainties and variable backgrounds, and parametric residual policy to handle pose error and intricate contact dynamics effectively. Leveraging the skill graph, our method supports knowledge-informed learning of semi-supervised learning for object detection and classroom-to-real reinforcement learning for fine manipulation. Simulation experiments on a gear-assembly task have demonstrated that the skill-graph-enabled coarse-operation planning and visual attention are essential for efficient learning and robust manipulation, showing substantial improvements of 13$\%$ in success rate and 15.4$\%$ in number of completion steps over competing methods. Real-world experiments further validate that our system is highly effective for robotic assembly in semi-structured environments. 
%This advancement is expected to significantly reduce the costs associated with policy design and environment customization, thereby enhancing the scalability and practicality of robotic assembly systems in flexible manufacturing contexts.
\end{abstract}
\begin{IEEEkeywords}
Robotic assembly, Semi-structured environment, Object detection, Semi-supervised learning, Residual reinforcement learning.
\end{IEEEkeywords}

\section{Introduction}
\label{sec:introduction}
\IEEEPARstart{T}{he} flexible manufacturing systems aim to swiftly adapt to market demands and individual customer requirements, facilitating a quick and cost-effective response to new tasks \cite{perzylo2019smeroboticsb1, hughes2020flexibleb2}. In contemporary industrial robotics, flexibility is primarily achieved through automated end-effector changes, efficient robot programming, and the utilization of component-specific fixtures \cite{Dharmara2018RoboticAO}. In the realm of low-volume batch production, robotic assembly systems tailored for flexible manufacturing must handle objects that are randomly positioned and unsecured by fixtures, thereby enhancing flexibility and adaptability across various product types in hardware \cite{nottensteiner2021towardsb3, suarez2018canb4}. While this less structured approach reduces the need for developing specific fixtures, saving both time and costs, it introduces significant software challenges, particularly the need for precise identification of randomly located objects within a defined workspace and the accurate control of force during precision assembly tasks. 

\begin{figure}[!t]
\centerline{\includegraphics[width=\columnwidth]{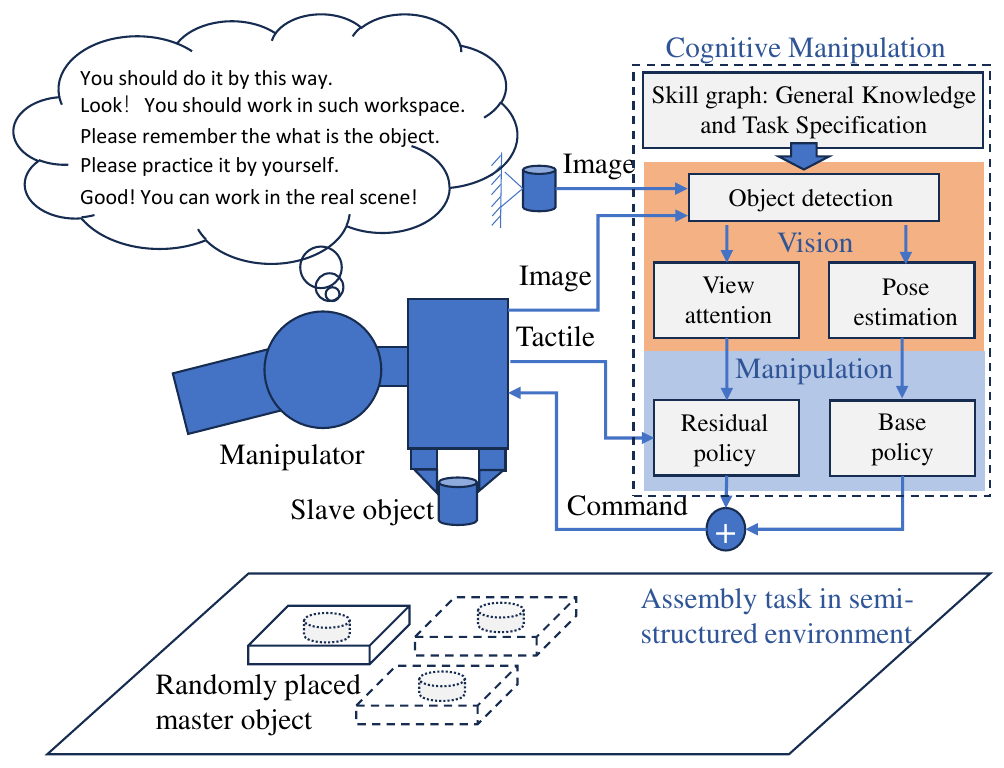}}
\caption{Robotic assembly tasks in a semi-structured environment and cognitive manipulation.}
\label{fig1}
\end{figure}

Human operators, leveraging task-specific knowledge and prior experience, intuitively manage these complexities to execute high-precision assembly tasks in such environments. Conversely, designing or learning an efficient and robust policy that enables robots to perform similarly in semi-structured environments poses a considerable challenge. 
%Existing robust assembly policies are generally categorized into multi-stage methods and end-to-end methods. 
Early solutions involved multi-stage methods that utilized vision systems to estimate pose errors and guide robots via visual servoing, complemented by force/torque-based algorithms to correct both visual and positional inaccuracies during the insertion process \cite{chen2007integratedb5}.
%To enhance robustness against variations in surface geometry and lighting conditions, and to simplify feature design, deep learning-based methods have been employed. These methods use object detection algorithms to identify relevant regions and estimate poses for visual servoing \cite{haugaard2021fastb6}. 
More recently, deep reinforcement learning (DRL) has emerged as a promising alternative, formulating effective policies through trial-and-error learning without relying on precise modeling and sensors \cite{luo2021robustb7}. 
However, these methods typically require extensive empirical knowledge or substantial training data, which can be time-consuming and labor-intensive, involving tasks such as image labeling, parameter tuning, or costly interactions.
%most of the existing model-based and learning-based systems are designed for stationary objects in a fixed context with satisfactory performance and cannot achieve good performance in a variable context environment. 

To address this issue, recent advancements in robotic manipulation have focused on the integration of multi-stage methods and deep reinforcement learning to  obtain robust and efficient policies\cite{shi2021proactiveb17, Zhao2023ALT}. The neural-symbolic framework integrates the convenience of traditional methods with the flexibility of RL, compensating for the inaccuracy of traditional positioning methods and the low efficiency of RL. 
Despite these advancements, significant challenges remain in utilizing the advantages of various modules for fixture-less precise assembly tasks. \textcolor{blue}{The primary issue is that singular visual or operational representations fail to provide a comprehensive understanding of the robots, tasks, and environments necessary for robot learning to handle the large position uncertainty and complex contact dynamics \cite{Stevsic2020LearningTA, Yu2024ManiPoseAC}. Additionally, both hand-designed and learning-based prior visual and operational representations often require substantial engineering efforts or extensive training data, complicating their application in real-world scenarios \cite{Köhler2023FewSOD}.} 

A key differentiator between human and non-human cognition is the capacity for structured knowledge representation, which has proven essential in addressing these challenges \cite{Zhang_2023}. Such representations linked with visual \cite{demura2018pickingb8, yen2020learningb9, hsieh2022deepb9-2} or operational priors \cite{Johannsmeier2018AFF, Li2019AFM, Liu2023HierarchicalRL} and learning process guided by knowledge \cite{Sun2024Dual}, have successfully captured a wide range of human cognitive processes, enabling efficient autonomous learning with minimal supervisory intervention \cite{Yang2019RobotLS, Rayyes2023InterestDrivenEW}. Drawing inspiration from the human approach to learning and manipulation, this thesis introduces the Cognitive Manipulation Method for Robotic Assembly in Semi-Structured Environments (CM4RASSE). This innovative method utilizes graph-based structural prior knowledge to integrate learning-based object detection and fine manipulation models into a cohesive modular policy, promoting self-directed learning with minimal human oversight. 

The proposed approach begins by establishing a skill graph that consolidates spatial, temporal, and causal information, creating a structured and flexible cognitive manipulation architecture by integrating multiple modules. This framework complements the general object detection model as a rich visual representation for fine manipulation by providing operationally relevant positional and visual attention information. Additionally, a rich operational representation based on skill graphs and planning enables transitioning from global observation and coarse operation to local observation and fine operation to address the challenges of assembly tasks in semi-structured environments. 
The proposed knowledge-informed developmental training method mirrors the human cognitive process of observing before acting and mastering skills in controlled settings before tackling real-world scenarios. 
Initially, we employ the skill graph and to collect diverse samples from pick-and-place interactions and minimal manual labeling for calibration and automated
labeling, facilitating semi-supervised object detection learning and cost-effective hand-eye-task calibration. This phase results in a rich vision representation which is robust to position uncertainty and variable backgrounds. 
Subsequently, the skill graph integrates visual perception with coarse operation planning, providing structured data that includes spatial locations and task-focused perspectives, enabling the agent to learn a residual policy for manipulation skills instead of learning from scratch. Moreover, this structured approach facilitates the transfer of learned skills to semi-structured environments, where variability and unpredictability are more pronounced. 
The efficacy of CM4RASSE is first studied in simulation and subsequently evaluated on a real robot through its application in a high-precision, jigless peg-in-hole and gear assembly task. The results show that the vision and manipulation model can be efficiently learned in new tasks with prior knowledge guidance, and the integrated policy performs effectively in the semi-structured environment.

This work presents a cognitive manipulation method, leveraging a skill graph to integrate learning-based visual perception and fine manipulation to facilitate efficient and effective learning of new tasks. Our primary contributions are as follows: 
\begin{enumerate}
\item{A Novel neural-symbolic framework: We have developed a skill graph that serves as common sense to integrate various modules, thereby handling the large position uncertainty and complex contact dynamics.}

\item{Semi-Supervised visual representation learning: We leverage the skill graph to collect a diverse array of samples and minimal manual labeling for Cost-Effective Hand-Eye-Task calibration and automated labeling, promoting semi-supervised learning of  Object Detection.}  %This strategy significantly reduces the cost and labor typically associated with extensive manual data labeling and calibration for manipulation. 

\item{Classroom-to-Real Residual Reinforcement Learning: Our skill graph, combining visual perception with trajectory planning, supports the learning of fine manipulation policies within a structured environment and enabling a seamless transition of acquired skills to a semi-structured environment, effectively navigating the inherent challenges of precise assembly tasks.} 

\item{Comparative and Comprehensive Studies: We conduct comparative and comprehensive studies to assess the effectiveness of each component and integrated policy in terms of learning efficiency and assembly performance. } %These studies are crucial for demonstrating the practicality and feasibility of our approach in executing precision assembly tasks within semi-structured environments.
\end{enumerate}

\section{Related work}
Robotic systems require precise state feedback and sophisticated control policies to effectively address specific tasks in semi-structured environments. \textcolor{blue}{This section reviews the cutting-edge methodologies and significant advancements in robotic assembly within such environments, highlighting the visual representation and operational representation for robot learning and concepts derived from human cognitive systems.}

\subsection{Robotic Assembly in Semi-structured Environments}
Multi-stage methods have been pivotal in precision assembly tasks, employing an integrated robotic system across three distinct phases \cite{hughes2020flexibleb2}: 1) Initial Approach: Utilizing an eye-to-hand camera, the system employs position-based visual servoing (PBVS) to navigate towards the master object. 2) Alignment: A force/torque-based local search method corrects alignment errors to ensure precise fitting. 3) Insertion Execution: Discrepancies in position and orientation are rectified using a force/torque control algorithm, ensuring successful component insertion. To enhance the efficiency of local search for assembling components with complex geometries, the part's geometry itself guides the alignment process through image-based visual servoing (IBVS) with an eye-in-hand camera \cite{song2014automatedb12}. Hybrid strategies that combine both eye-to-hand and eye-in-hand cameras merge the benefits of different visual servoing techniques, providing comprehensive visual cues that maintain target visibility throughout the operation \cite{krishnan2020performanceb13, peng2020comparingb14}. However, the success of these systems heavily depends on the precise selection of features and the strategic design of control methodologies, necessitating meticulous calibration and parameter adjustments to minimize errors and ensure operational stability.

Innovative approaches have been introduced to bolster robustness against variations in surface geometry and lighting conditions and simplify vision, calibration, and control processes. For instance, Haugaard et al. \cite{haugaard2021fastb6} introduced a deep learning approach for pin and hole point estimation in multi-camera setups, facilitating visual servoing for initial alignment. Mou et al. \cite{mou2022poseb15} devised a technique for more precise and efficient position estimation of manipulated connectors, leveraging YOLO-based relevant region detection. \cite{2020LearningTA} propose a 6D pose estimation of template geometries, to which manipulation objects should be connected. To diminish design complexity and boost optimal policy, reinforcement learning (RL) \cite{luo2021robustb7} presents an alternative by leveraging trial-and-error learning over precise modeling. \textcolor{blue}{To heighten approach robustness amidst environmental variations, spatial attention point network models \cite{Yasutomi2023VisualSA} have been introduced, employing visual attention to extract pertinent image features for motion controllers and utilizing offline training to enhance sample efficiency. However, it is expensive to collect sufficient experience data from real-world scenarios. Furthermore, the low clearance and contact dynamics in precision assembly tasks complicate demonstrations \cite{Schoettler2019DeepRL}, simulation-to-real transfer\cite{Wang2021AlignmentMO}, and offline training.} 

Several studies have also leveraged model-based and learning-based modules to construct robust and efficient policies. Based on learning in multi-stage, Lee M A et al. \cite{lee2020guidedb16} employed vision-based uncertainty estimation to differentiate between free-space and contact-rich regions, applying model-based methods in free-space for minimal environmental interaction and RL techniques to navigate inaccuracies in perception/action pipeline. Zhao et al. \cite{Zhao2023ALT} proposed a fine positioning policy learned by DRL under an eye-in-hand camera view with a traditional coarse positioning method and impedance control. Based on the residual learning, Shi Y et al. \cite{shi2021proactiveb17} combined an eye-in-hand vision-based fixed policy with a contact force-based parametric policy to enhance the robustness and efficiency of the RL algorithm. Besides the force-based trajectory generators, \cite{Ahn2023RoboticAS} introduced an image-based trajectory generator trained by DRL to enable a robot to adapt to assembly parts with different shapes. Similarly, \cite{Zhang2024ARR} proposed a residual high-level visual policy to determine the robot pose increment in Cartesian space through deep RL. 
However, these methods with direct bonding cannot solve the large position uncertainty and complex contact dynamics simultaneously and it is difficult to adapt to new tasks quickly because they do not fully utilize the advantages of various modules. 

%This study concentrates on precise assembly in semi-structured environments, confronting challenges like significant positional uncertainty, variable backgrounds, and complex contact dynamics. The cognitive manipulation flow first from global observation and coarse operation to local observation and fine operation by knowledge-based skill, integrating data from two cameras and force feedback. This combination offers coarse motion and observation guidance via two YOLO-based detectors and partial model-based planning for adaptability and efficiency in fluctuating environments. A DRL-based residual policy merges focused views and tactile feedback to correct detector inaccuracies and enable adaptable force control in contact-rich manipulations. 

%In addition, the proposed method incorporates a hybrid model and learning-based approach in perception and manipulation, trained separately with prior knowledge guidance to increase efficiency and robustness.

\subsection{Visual Representation for Robot Learning}
The integration of prior visual models into the RL framework has shown promise in enhancing learning efficiency and generalization in unstructured settings by detection \cite{demura2018pickingb8}, pose estimation \cite{chen2021deepb10, hsieh2022deepb9-2}, visual affordances \cite{yen2020learningb9}. Unsupervised \cite{Zhang_2023}, self-supervised learning \cite{Jin2023VisionforcefusedCL} and hybrid observation-synthesis \cite{Chen2023Autotr} has been applied to learn the prior visual models for different robotic skills. Specifically for grasping, \cite{borja2022affordanceb11} propose self-supervised visual affordance models that are grounded in real human behavior from teleoperated play data, driving the model-based planner to the vicinity of afforded regions and guiding a local grasping RL policy to favor the same object regions favored by people. Building on this prior work, this study introduces a semi-supervised visual representation to provides structured information including spatial location and task attention information for assembly skill learning.
\subsection{Operational Representation for Robot Learning}
Operational representation can reduce the complexity of the solution space of a given manipulation problem by applying a well-designed but still flexible structure, such as formal method for task and domain-specific knowledge \cite{Li2019AFM}, stochastic graph \cite{Xiong2016RobotLW}, switching functions \cite{Yu2022adap}, manipulation primitives \cite{Tavassoli2022LearningSF}, graph-based skill formalism \cite{Johannsmeier2018AFF, Liu2023HierarchicalRL}. Especially, \cite{Liu2023HierarchicalRL} uses temporal abstraction and task decomposition as the higher-level policy in the hierarchical reinforcement learning method to reduce problem complexity. Based on this work, we further extend the skill graph to fixture-less assembly tasks by integrating object detection, coarse operation planning, and residual fine manipulation policy.

\subsection{Cognitive Systems and Learning Mechanisms}
Research in cognitive robotics aims to emulate human intelligence, paving the way for the development of human-level artificial intelligence by cognitive architectures leveraging core capabilities such as sensing, cognition, learning, and control \cite{Li2019Combined}. The existing theories offer crucial insights for creating foundational elements and learning strategies for cognitive systems, such as hybrid neural-symbolic models \cite{Sun2024Dual} and top-down learning \cite{Xie_Jin_2018}.
% \cite{Asada2009CDR, Ersen2017CERM, Kotseruba_Tsotsos_2020}
%Learning based on prior knowledge as common sense is an important way for humans to achieve efficient learning \cite{Xie_Jin_2018, Zheng_Pai_2021}, which has inspired knowledge-based efficient and autonomous learning methods for robots. 
%The existing theories offer crucial insights for creating the foundational elements and learning strategies for cognitive systems, facilitating knowledge-based learning \cite{Hakimzadeh_Xue_Setoodeh_2021}. Schema and behaviorism theories play a significant role, with schemas focusing on the mind's building blocks and behaviorism emphasizing the influence of environmental factors on behavior. Additionally, the dichotomy between top-down and bottom-up learning, which involves acquiring explicit and implicit knowledge in varying sequences, is noteworthy.
Especially, \cite{Lopez-Juarez_2005} adopted a connectionist-based approach for object recognition and compliant motion learning based on adaptive resonance theory (ART), aiming to design robotic agents for assembly tasks. This study employs a skill graph to integrate the neural models, which enable human-like operation and learning. 

\section{Problem Statement}
\begin{figure}[!t]
\centerline{\includegraphics[width=1.0\columnwidth]{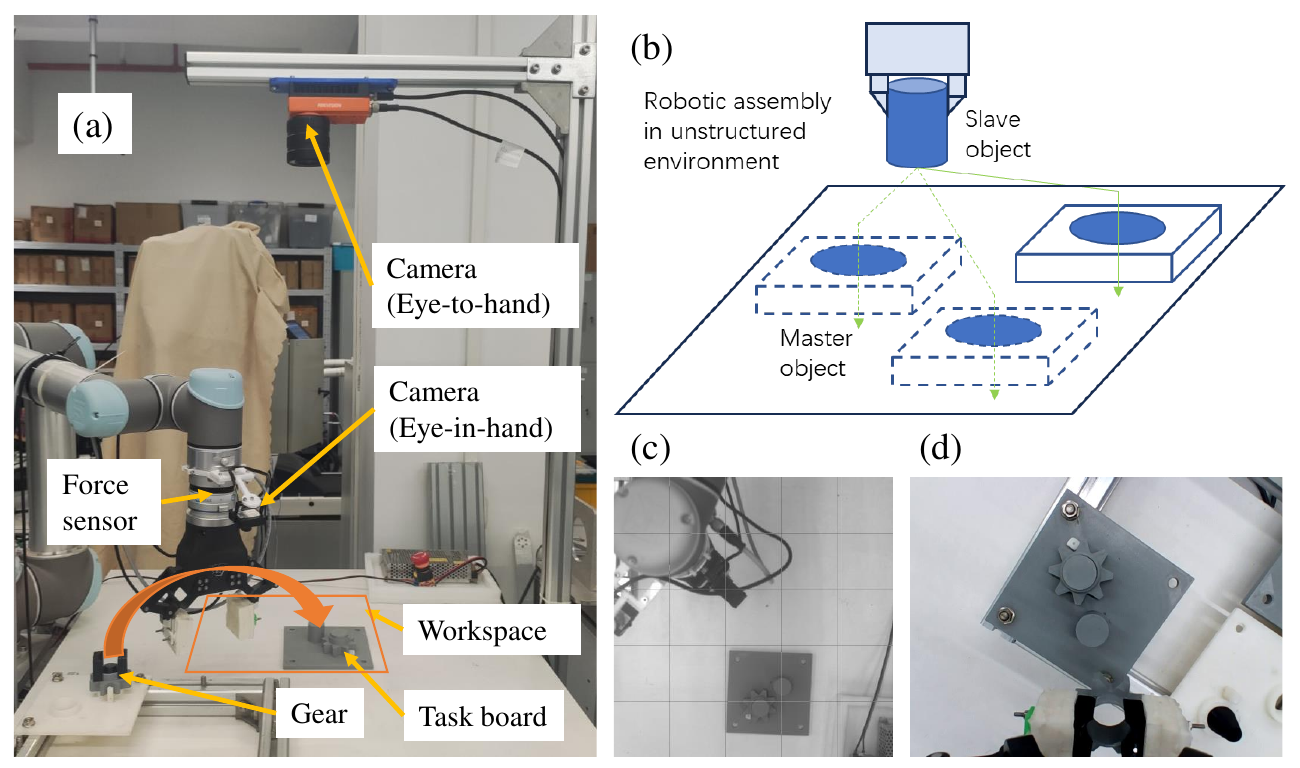}}
\caption{Robotic assembly in a semi-structured environment. (a) This work considers assembly tasks in a semi-structured deployment environment for simulating flexible manufacturing scenarios with a task board randomly placed in a predefined workspace. (b) We focus on the subtask of the robot assembly task, which uses vision and haptics to assemble the grasping slave object into the master object. (c) The eye-to-hand camera can see the whole workspace, but the assembly objects are hidden by the robot during the contact-rich manipulation phase. (d) The eye-in-hand camera can see the assembly objects clearly, but the limited field of view does not allow a continuous view of the entire work area.}
\label{Fig 5-1}
\end{figure}

This work focuses on the ability to locate the master object and insert the slave object where the master object is randomly positioned within the workspace, as shown in Fig. \ref{Fig 5-1}. %We undertake a gear assembly task characterized by uncertain positioning and contact dynamics, executed by a UR5 robot equipped with a two-finger gripper. The inclusion of additional sensors, such as two cameras and a force sensor, supplies multimodal insights into both the task and its surroundings. Some degree of predictability and regularity about robot, task, and environment is available for control policy development. 

The task can be formulated as a Markov Decision Process (MDP) with a transition function $P (S_{t+1} | S_t, A_t)$, which is a probability distribution over the next state $S_{t+1}$ conditioned on the execution of a certain action $A_t$ in the current state $S_t$. We want to find a policy $\pi (A_t | S_t)$ that dictates the probability over actions conditioned on a given state. The complexity of the transition function $P$ is determined by the degree of the environment, which directly affects the difficulty of designing or learning the policy $\pi$. Unlike the environment in which humans live, more structured industrial scenarios can provide more prior knowledge for the learning process \cite{Chen2007IntegratedRS, von2022uncertaintyb19, lee2020guidedb16, Eer2023GuidedRL}. Therefore, the assumption of known partial knowledge of the state $S_t$ and transition function $P$ is exploited as follows: 
\begin{enumerate}
\item{Semi-structured industrial scenarios, such as flat workbenches and limited work areas, add constraints to the robot's behavior and can also simplify the robot's perception and operation requirements. \textcolor{blue}{In this work, the dimensions of the master object's pose are categorized into constrained and unconstrained parts. The constrained segments of $z$, $rx$, and $ry$ are restricted within the workspace, influenced by the task and workspace shape. Conversely, the unconstrained segment of $x$, $y$ and $rz$ varies within a specific range ($S_{uncon}: [(x_{min}, x_{max}), (y_{min}, y_{max}), (rz_{min}, rz_{max})]$).} }
\item{General knowledge about manipulation tasks can guide strategy design and learning, considering the spatial separation between the master and slave objects. \textcolor{blue}{The manipulation process can be divided into contact-free $S_{cf}$ and contact-rich $S_{cr}$ regions based on task geometry, with attention to uncertainties in pose estimation $E_r$ and contact dynamics $F^{max}$ in contact-rich $S_{cr}$ regions for precise assembly tasks. Humans utilize global and local fields of view sequentially to enhance fine manipulation tasks, addressing the constraints of a single camera. } }
\item{The geometric parameters and even the forward and inverse kinematics of the robot are often available from the robot supplier. \textcolor{blue}{It allows us to design and learn the manipulation policy in task space. We can also use it to obtain geometry information of tools, platforms, and tasks by demonstration.}}
\end{enumerate}

%1)半结构化的工业场景，例如平整的工作台、有限的工作区域等，给机器人的行为添加了约束，同时也可以简化机器人感知能力的需求。2)操作任务的通用知识可以为策略设计和学习提供指导。3)机器人的几何参数甚至正逆运动学往往是可以从机器人供应商那里获得的。

The challenge is to incorporate general prior knowledge and a learning-based model to address task-specific uncertainty. This work aims to propose a neural-symbolic cognitive manipulation method for assembly skill learning, enabling the utilization of prior knowledge to train a visual representation and fine manipulation policy to handle the uncertainties of robot, environment and tasks.
%allows us to utilize prior knowledge to train a visual representation and subsequently apply these priors to train the fine manipulation policy. It is promising for practical application to learning a robust policy without known geometrical-physical information of tools, platforms, and tasks in a controllable environment, which can then be transferred to unstructured settings.

\section{Method}
\begin{figure}[!t]
\centerline{\includegraphics[width=\columnwidth]{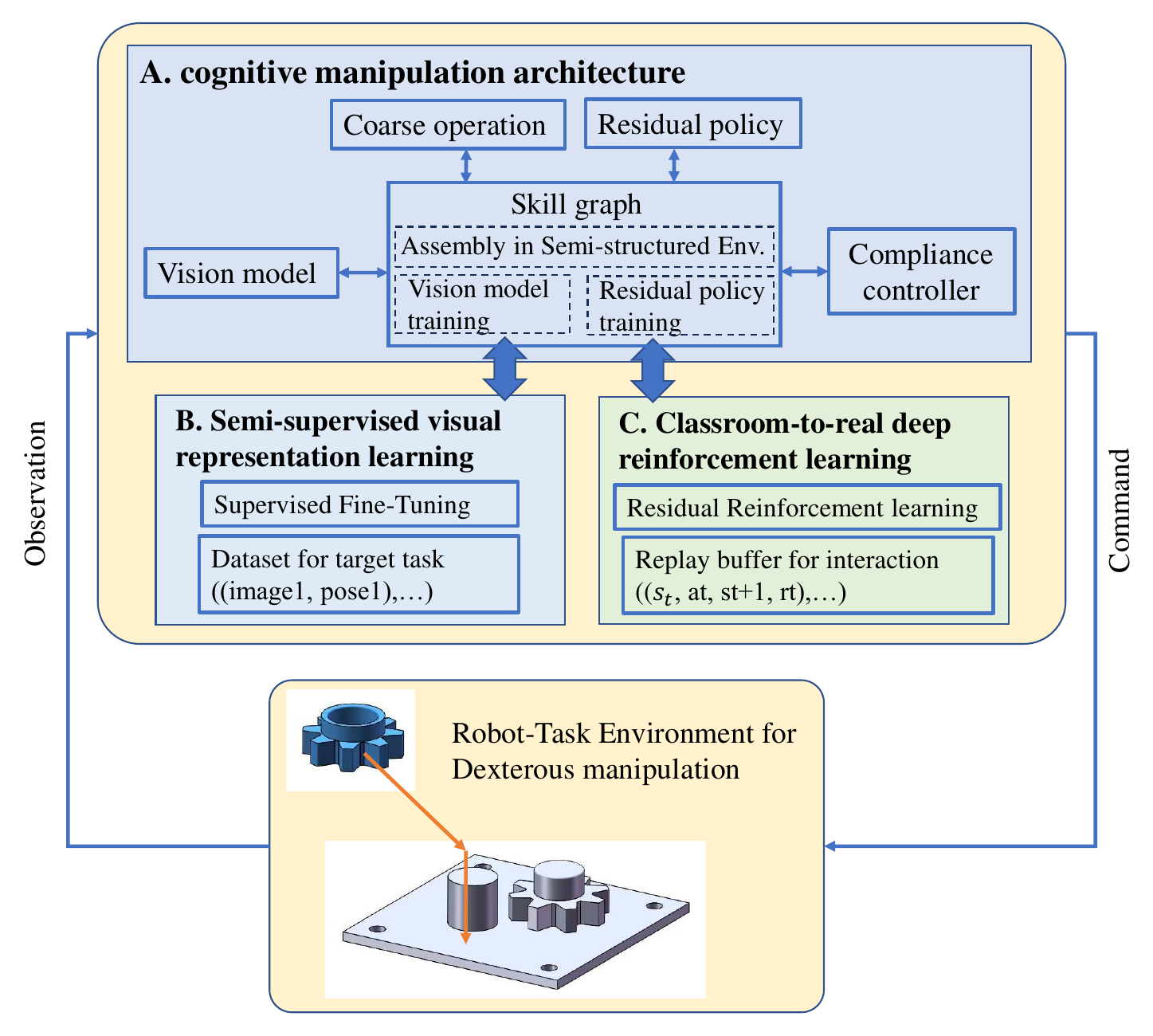}}
\caption{The cognitive manipulation architecture with semi-supervised visual representation Learning and classroom-to-real reinforcement learning.}
\label{Fig 5-2}
\end{figure}

This work introduces a novel cognitive manipulation method for solving assembly tasks in semi-structured environments, as shown in Fig. \ref{Fig 5-2}. \textcolor{blue}{Central to our approach is the skill graph, which orchestrates multiple modules within a mixed-strategy framework, driving three key modes of operation: manipulation in semi-structured environments, vision model training, and fine manipulation training.} 
%The method utilizes a partial geometric model of the task and environment to enhance object detection, facilitating the accurate estimation of the target assembly pose for the robot end-effector from a global perspective while generating focused task attention in the local view. Flexible coarse operation planning utilizes estimated pose to generate efficient free-space motions and ensure safe operations in contact-rich regions. In the subsequent phase, a focused-view residual policy is activated to compensate for the uncertainties associated with pose localization and contact dynamics. 
%To achieve efficient, task-tailored learning of cognitive manipulations, the skill graph first schedules coarse operations to collect passive observation data. This data is then semi-automatically labeled to enable semi-supervised learning of object recognition models. Under the guidance of coarse operations and task attention view, the residual policy for contact-rich manipulation is then learned by reinforcement learning (RL) in a structured environment. \textcolor{blue}{Training data storage acts as a learning memory, while model storage acts as a working memory, highlighting the dynamic interplay between data acquisition and operational execution.}
This section presented the proposed methodology in three parts: 1) Cognitive Manipulation Architecture: This component introduces a neural-symbolic framework that integrates multimodal and scalable information through a combination of model-based and learning-based methods. 
%This architecture is designed to adaptively respond to the complexities of semi-structured environments. 
2) Semi-supervised Visual Representation learning: This stage outlines a cost-effective method for hand-eye-task calibration and object detection training, enhancing the visual representation capabilities essential for precise manipulation. 
3) Classroom-to-Real Residual Reinforcement Learning: The final part involves training the residual policy within a specially designed classroom environment and task execution in semi-structured settings. %This curriculum-based approach progressively instills complex manipulation skills necessary for successful task execution in semi-structured settings.

%论文提出了一种知识驱动的学习与操作框架。knowledge base作为通用知识，由人类示教获得几何信息实现粗操作task specification，其驱动混合策略的多个模块实现操作，避免遮挡等问题，驱动采样过程，实现高效和有效的学习。其作为thinking驱动操作和学习。训练过程中的数据是学习记忆，操作过程中的模型存储是工作记忆。
%the knowledge base is taught by human beings to obtain geometric information to achieve coarse operation task specification

\subsection{Cognitive Manipulation Architecture}
\begin{figure}[!t]
\centerline{\includegraphics[width=\columnwidth]{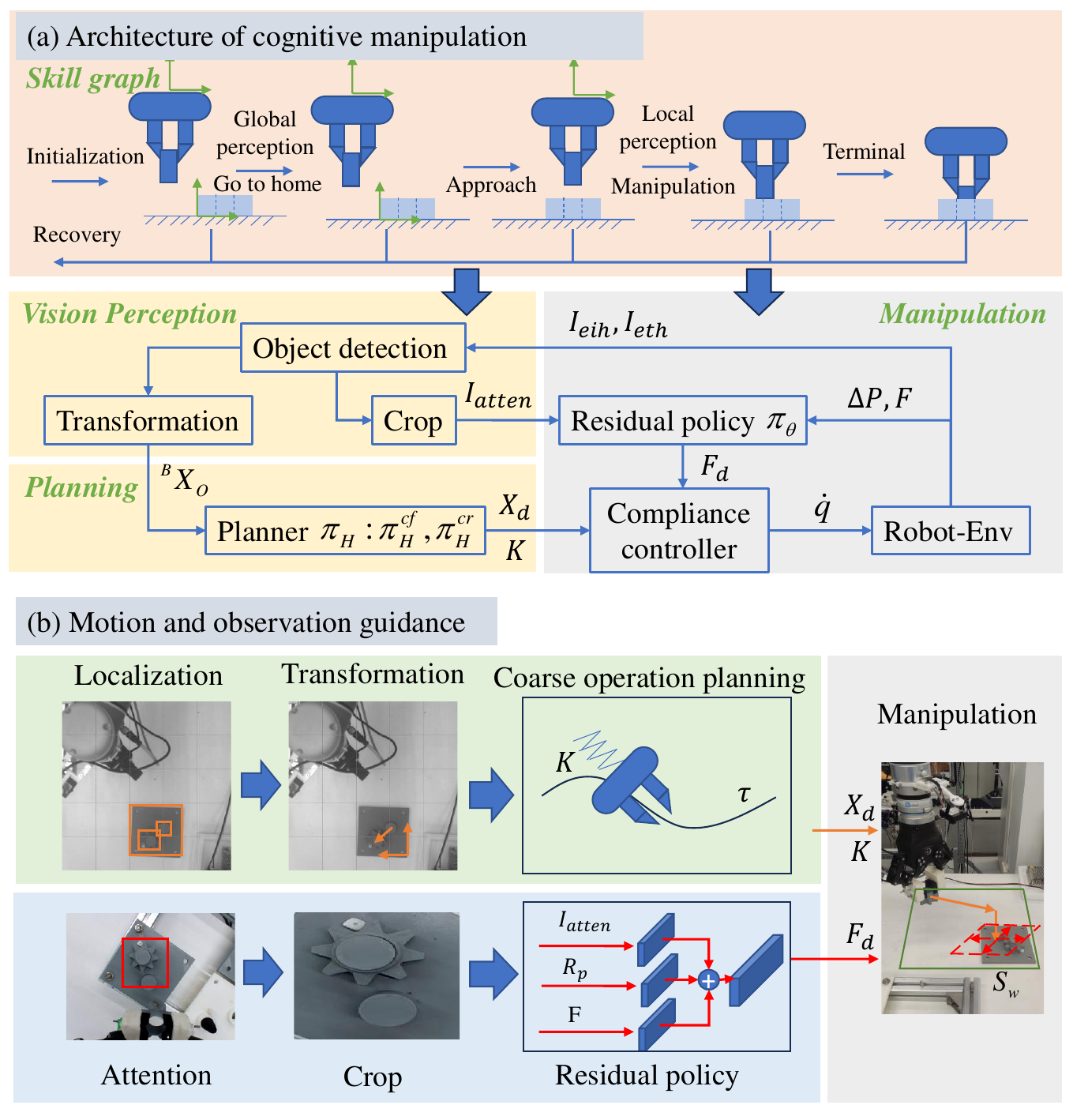}}
\caption{Components of cognitive manipulation for precise assembly tasks in a semi-structured environment. A skill graph divides the manipulation into multiple stages solved with multiple modules. The object detection estimates rough location ${}^BX_O$ and provides task-related features $I_{atten}$. The planner generates trajectory and stiffness as coarse operation policy $\pi_H$. In the free space, the impedance-controlled trajectory $\pi_H^{cf}$ moves the robot to the contact-rich $S_{cr}$ regions. Once the task is in the region, $\pi_H^{cr}$ is switched and the residual policy $\pi_\theta$ is enabled for fine alignment and contact dynamics in contact-rich manipulation.}
\label{Fig 5-3}
\end{figure}

The cognitive manipulation architecture is designed to enhance the effective contact-rich manipulation in semi-structured environments. It leverages a skill graph to integrate multiple modules including an object detection module to manage positional uncertainties, a model-based system for trajectory planning and compliance settings, and a residual policy for managing pose error and complex contact dynamics. These components work together to support effective manipulation control, as depicted in Fig. \ref{Fig 5-3} and Alg. \ref{alg1}. 
%We commence by introducing a skill graph and the uncertainty inherent in the available partial model utilized for manipulation. Subsequently, we introduce an object detection module tailored to address positional uncertainties in dimensions that lack fixture, while concurrently generating visual attention directed towards the task at hand. We then detail a flexible, model-based approach for planning trajectories and compliance parameters essential for effective motion control during manipulation. The section culminates with the introduction of a residual policy, enhanced with visual attention, designed to refine the initial planned policy through precise localization adjustments and management of complex contact dynamics.

\subsubsection{Skill Graph Based on General Knowledge and Task Specification}
\begin{figure}[!t]
\centerline{\includegraphics[width=0.9\columnwidth]{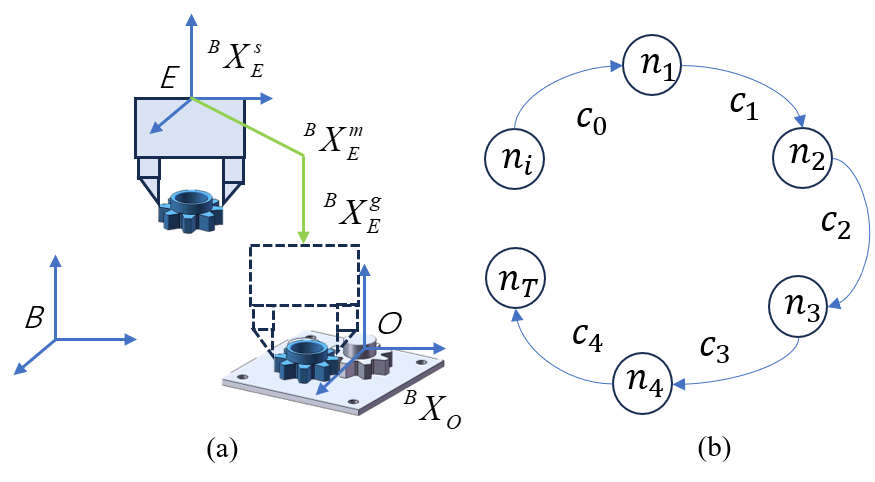}}
\caption{The geometric information and temporal logic of manipulation.}
\label{Fig 5-4}
\end{figure}

\textcolor{blue}{The abstract expert knowledge of assembly tasks in semi-structured environments can be harnessed to segment the manipulation process into distinct stages and components, as depicted in Fig. \ref{Fig 5-4}. The skill graph integrate symbolic and subsymbolic representations according to general versus task-specific knowledge and degrees of accessibility for a well-designed yet flexible structure.} 
  
%with knowledge-informed robot learning. This integration facilitates the segmentation of the manipulation process into distinct stages and establishes a foundational policy for perception and operations, thereby guiding the agent in acquiring task-specific skills.
% This is achieved through

We define general knowledge using a partial model that encapsulates spatial, temporal, and causal information. Initially, we consider the spatial information concerning the end-effector (EE) pose in the manipulation, which includes the home position ${}^BX_E^s$, assembly bottleneck pose ${}^BX_E^{m}$, and assembly goal pose ${}^BX_E^g$ within the robot base frame. Temporally, the process is segmented into four stages: reaching ${}^BX_E^s$ for global perception to estimate the ${}^BX_E^g$, planning a coarse operation from ${}^BX_E^s$ to ${}^BX_E^{m}$ and then to ${}^BX_E^g$, executing the coarse operation to reach ${}^BX_E^{m}$, and performing the fine operation for insertion at ${}^BX_E^g$. Causal transition conditions between these stages are defined based on the positional relationships and contact states between the peg and hole. This partial model is illustrated in Eqn. \eqref{eqA-1} and Fig. \ref{Fig 5-4}. Although the introduction of sequential logic and causality is crucial for operating in semi-structured environments—enhancing safety and reducing learning costs due to potential interference from other agents (robots or humans), this paper primarily focuses on the sequential logic and causal enhancement of robot learning methods, while not extensively addressing multiple exceptional states and their management strategies.
%the home position of end-effector(EE) ${}^BX_E^s$ and master object position ${}^BX_O$ in robot base frame and bottleneck pose of EE ${}^OX_E^{m}$ and goal pose ${}^OX_E^g$ in task frame. So the bottleneck pose of EE ${}^BX_E^{m}$ and goal pose ${}^BX_E^g$ in robot base frame can be obtained from the master object position ${}^BX_O$.  For example, the mutual complement of global and local environment awareness, the combination of global operation and local operation, task status judgment and security judgment

\begin{equation}\label{eqA-1}
\begin{cases}
n_1: f_{gp}(\theta_1; \Omega_1) & c_1: ^BX_O \in S_{uncon} \\ 
n_2: f_{pl}(\theta_2; \Omega_2) & {c_2: done} \\
n_3: f_{cf}(\theta_3; \Omega_3)  & c_3: X - {}^BX_E^{m} \in E_{th} \\
n_4: f_{cr}(\theta_4; \Omega_4)  & c_4: X - {}^BX_E^g \in E_{th} \text{\&} F_z > F_z^{max}
\end{cases}
\end{equation}

%the temporal logic is expressed with four nodes. Initial node: go to the home position, $n_1:$ calculate the location of the robot end-effector and master object, $n_2:$ move to the bottleneck pose from current pose, $n_3:$ perform contact-rich manipulation for insertion to , Terminal node: go to the home position. Finally, causality is defined as transitions, transforming an input fluent to an output fluent by using an action from temporal logic. Furthermore, the additional transition is activated by an error condition C err and leads to a recovery node. Note that we do not elaborate on the causality and recovery node, since it is clearly out of the scope of this work. We only use the distance relative to the spatial position to convert between the temporal logic nodes. 
%Secondly, the temporal logic is expressed with four nodes. $n_1:$ go to the home position, $n_2:$ calculate the location of the robot end-effector and master object, $n_3:$ move to the bottleneck pose, $n_4:$ check the master object pose, $n_5:$ perform contact-rich manipulation, then open the gripper, then go to the home position.

\textcolor{blue}{Despite the generic nature of temporal and causal information in assembly tasks, task-specific information remains essential. For the manipulation process, the bottleneck poses ${}^BX_E^{m}$ and goal pose ${}^BX_E^g$ of EE are determined by the task's geometry and assembly relationships, master object pose, grasp pose for the slave object, and the tool center position (TCP) offset. Object-Embodiment-Centric (OEC) geometry representation is derived from demonstrations to enable direct prediction of key waypoints in the operational process through the master object pose estimation, as outlined in our prior work \cite{Wang2024OEC}. The OEC representation selects a grasping point on the task board as the coordinate origin ${}^BX_O$ and extrapolates the bottleneck pose of EE ${}^BX_E^{m}$ and goal pose ${}^BX_E^g$ in the robot base frame from the teaching, then transform these to ${}^OX_E^{m}$ and ${}^OX_E^g$ in the task frame.  
In the semi-structured environment, the master object is randomly placed within a workspace of range $S_{uncon}$, as specified by a human operator. The home position of the EE ${}^BX_E^s$ is strategically set outside this region to prevent occlusion of the eye-to-hand camera's field of view. With partial constraints of the workspace, the pose can be determined by estimating the $x$, $y$, and $rz$ dimensions. The uncertainties introduced by the pose estimation and demonstration system, as well as contact dynamics, are also considered for precise contact-rich task execution. 
In conclusion, employing the general knowledge base, both planning and learning-based methods are utilized for developing a complex policy, which includes object detection for pose estimation and task-related visual information extraction, spatially dependent trajectory planning as the basic strategy, and task-specific residual strategies for handling uncertainties. A graph structure is devised to reflect the required sequence of motions and modules necessary to complete the task, as illustrated in Fig. \ref{Fig 5-3}. } 

%The constrained parts $P_{con}$, range of unconstrained parts $S_{uncon}$, the relative pose of the end-effector to the task board frame $\Delta P_g$ and $\Delta P_p$ can be given by an expert. 
%In the $S_{cr}$, the relative assembly target pose $\Delta T_g$ is determined by the assembly relationship of the task and the pre-assembly pose $\Delta T_p$ is defined by the geometry of task. The relative pose of end-effector $\Delta P_g$ and $\Delta P_p$ can be determined by the relative grasp pose and the tool center position(TCP) offset. 

\subsubsection{Compliance Controller}
\textcolor{blue}{Employing the virtual force-driven spring-mass-damping modal and robot kinematics, we utilize a modified Cartesian parallel position and force controller as the low-level control for robot learning, generating velocity commands. The control law for joint velocities $\dot{q}$ is expressed as:}

\begin{equation}
\label{eq_5-4-A-2-1}
\overset{\cdot }{q} = \frac{J^{-1} M^{-1}}{s+M^{-1}B} \left [ K\left ( X_{d}-X \right ) -\left ( F_{d}-F \right )  \right ]
\end{equation}
where Jacobian matrix $J$ provides the relation between end-effector and joint velocities. Desired poses $X_d$ and forces $F_d$ govern the behavior, with the stiffness matrix $K$ balancing the six-dimensional tracking error for position/orientation and force/torque. The inertia matrix $M$ and damping matrix $B$ influence the response speed and stability.

\subsubsection{Object Detection for Pose Estimation and Task Attention}
Object detection is a popular algorithm used to locate objects in an image or video stream. It predicts multiple bounding boxes for objects in the image $I$, and each bounding box contains the predicted values for the object's position $(x, y)$, size $(w, h)$, confidence $c_{con}$, and category $c_{cate}$, as shown in \eqref{eq2}. With a pre-defined confidence level, the effective predicted bounding box for the object is selected.

\begin{equation}\label{eq2}
[c_{cate}, x, y, w, h, c_{con} ] = detect(I) 
\end{equation}

To cover the entire workspace and accurately detect the object of interest, we attach an eye-to-hand RGB camera at the top of the workspace to capture the 2D image $I_{eth}$, as shown in Fig. \ref{Fig 5-3} (b). An object detection-based coarse perception system generates one bounding box around the object of interest to obtain the location $(x_0, y_0)$ and two other bounding boxes around the predefined feature structures to obtain the location $(x_1, y_1)$ and $(x_2, y_2)$. According to the eye-to-hand transformation ${}^rT_c$, the estimated points $(x_{0}', y_{0}')$  are transformed to the robot frame as shown in \eqref{eq3}. Considering the partial pose information of the object, including $rx_{con}$, $ry_{con}$, and $z_{con}$, the pose ${}^BX_O$ can be determined as in \eqref{eq4}. \textcolor{blue}{Global perception is used in the first stage $n_1$ to determine whether the main assembly object is ready for the assembly operation on the one hand, and to provide location information for the assembly operation on the other hand.}

\begin{equation}\label{eq3}
(x_{i}', y_{i}') = tramsform (x_{i}, y_{i}), i=0,1,2
\end{equation}

\begin{equation}\label{eq4}
{}^BX_O = [x_{0}', y_{0}' , z_{con}, rx_{con}, ry_{con}, \arctan( \frac {y_{2}-y_{1}}{x_{2}-x_{1}} )]
\end{equation}

The second model uses the image $I_{eih}$ from an eye-in-hand camera to provide local task detection \textcolor{blue}{as attention}, enhancing the ability to differentiate tasks from the environment as indicated in Fig. \ref{Fig 5-3} (b). Object detection utilizes a bounding box as a region of interest (ROI) to identify specific structures crucial for vision-based precise alignment in assembly tasks. The work uses a simple attention strategy that utilizes this bounding box $(x_a, y_a, w_a, h_a)$ to crop and resize the task-related area from the input image $I_{eih}$ and generate an attention-guided observation $I_{atten}$ for fine manipulation, enabling the residual policy to concentrate on the specific structure amidst varying environments. 

\begin{equation}\label{eq5}
I_{atten} =crop((x_{a}, y_{a}, w_{a}, h_{a}), I_{eih}) 
\end{equation}

\subsubsection{Coarse Operation Planing}
With the partial model and coarse perception system, we can plan a coarse operation as the second stage $n_2$. We first obtain the assembly goal point ${}^BX_E^g$ and bottleneck pose ${}^BX_E^{m}$ with the estimated pose ${}^BX_O$ and the OEC geometry information of ${}^OX_E^{m}$ and ${}^OX_E^g$, which divide the operation into contact-free and contact-rich manipulation. 

The uncertainty due to pose estimation and compliance control can be ignored in the contact-free region $S_{cf}$. A fast min-jek trajectory $\tau_{cf}$ between the home point ${}^BX_E^s$ and the bottleneck pose ${}^BX_E^{m}$ can be generated. In addition, a high-stiffness $K_{cf}$ of compliance controller is used to ensure acceptable position tracking errors. The trajectory and stiffness provide coarse operation for the contact-free region, which can be defined as Eqn. \eqref{eq6}. 

\begin{equation}\label{eq6}
\pi ^ {cf}_H \sim (\tau _{cf} (t),K_{cf}) 
\end{equation}

However, the uncertainty cannot be ignored in the contact-rich region. A slow trajectory $\tau_{cr}$ and small stiffness are used in the contact-rich region $S_{cr}$. We define an exploration space $W$ to represent the offset range of a compliance robot disturbed by safe external forces $F_{max}$. It should cover the assembly depth and error range to ensure safe contact and effective error compensation, as shown in Eqn. \eqref{eq7}. Furthermore, the small stiffness matrix of compliance control $K_{cr}$ is obtained with the estimated exploration space $W$ and maximum contact force $F_{max}$, which can be defined as,

\begin{equation}\label{eq7}
\begin{aligned}
&W =  E_{r}  + abs( {}^BX_E^{m} -{}^BX_E^g )  \\
&K_{cr}  = F_{max}  \cdot  diag  (W)^ {-1}  \\
&\pi ^ {cr}_H \sim (\tau _{cr} (t),K_{cr})
\end{aligned}
\end{equation}

\subsubsection{\textcolor{blue}{Residual Policy for Fine manipulation}}
The manipulation is divided into two phases according to the planned coarse operation and carried out by the compliance controller in \eqref{eq_5-4-A-2-1}. \textcolor{blue}{In the third stage $n_3$, the end effector of the robot is moved from the home point ${}^BX_E^s$ to the bottleneck pose ${}^BX_E^{m}$ with a planned efficient policy $\pi ^ {cf}_H$. }

Since the contact-rich assembly manipulation in the fourth stage $n_4$ requires a higher level of accuracy than conventional robot and vision systems, the planned safe policy $\pi ^ {cf}_H$ is switched and the residual policy $\pi_{\theta}$ is enabled to refine the initial policy for precise localization and complex force control. In addition to guidance from a fixed policy, the residual policy also receives attentional observation guidance $I_{atten}$ from object detection. The tactile $F$ \textcolor{blue}{from the force sensor mounted on the wrist} and the relative pose $R_p = {}^BX_E - {}^BX_E^g$ of the end-effector serve as additional observations for contact dynamics. The residual policy generates the desired force and torque $F_d$ in Eqn. \eqref{eq8}, which together with $\pi ^ {cf}_H$ as input to the compliance controlle in Eqn. \eqref{eq_5-4-A-2-1}.

\begin{equation}\label{eq8}
F_d = \pi _ {\theta}  (I_{atten}, R_p, F)
\end{equation}

With the help of coarse operation planning, the learning of cognitive manipulation is carried out separately. The object detection models and hand-eye-task calibration are trained by semi-supervised visual representation learning in subsection B. The residual policy is trained by classroom-to-real residual reinforcement learning in subsection C.

\subsection{Semi-supervised Visual Representation Learning}
This section delves into training two object detection models and calibrating hand-eye-task relationships using collected samples based on the geometric model of a specific assembly task and the robot's kinematics. Our approach enables the gathering of a varied sample set through a carefully planned coarse operation in pick-and-place.  To address the challenges related to accurate data labeling, we suggest a streamlined calibration and labeling process that significantly reduces the engineering effort. Furthermore, fine-tuning from a pre-trained model is utilized to reduce the reliance on extensive sample volumes.

\begin{figure}[!t]
\centerline{\includegraphics[width=\columnwidth]{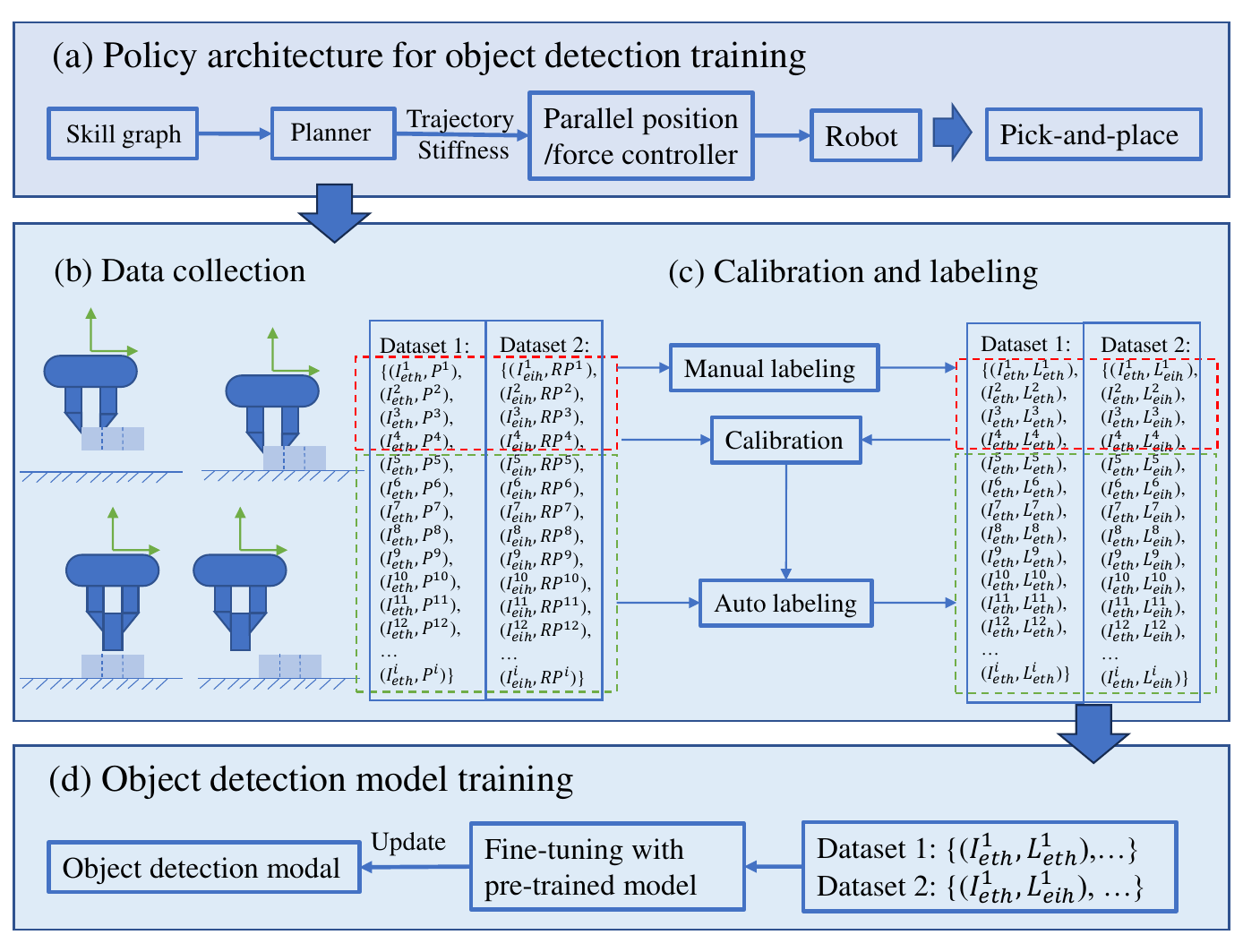}}
\caption{Semi-supervised Visual Representation Learning for object detection models and hand-eye-task calibration. (a) shows the policy architecture for embodied data collection for object detection model training. (b) shows the pick-and-place and sampling process designed based on the workspace and task information to collect diverse samples. Image and position information are recorded for calibration and model training. (c) shows the calibration and labeling process. (d) shows the model training process by fine-tuning with a pre-trained model.}
\label{fig4}
\end{figure}

\subsubsection{Data Collection via Coarse Operation}
The master object may appear at different points $P$ throughout the workspace $S_w$, requiring global localization, further global localization exists as possible relative pose error $R_P$ within the range $E_r$, requiring further local perception as visual attention for fine manipulation. To ensure data diversity for training the position and attention models, we first uniformly collected $m$ points within the workspace where the task board will appear during the real scene, denoted by $P = [x^{r}, y^{r}]$. Second, we used an offset to the bottleneck pose ${}^BX_E^{m}$ and generated $n$ points to cover the uncertainty space while avoiding collisions, defined them as $R_P = [\Delta x^{r}, \Delta y^{r}, \Delta z^{r}]$. The sampling points and poses are shown in Fig. \ref{fig4} (b). We collected eye-in-hand and eye-to-hand images, as well as the corresponding position and relative pose, using a hand-designed trajectory for \textcolor{blue}{data diversity}, as shown in Eqn. \eqref{eq9}.

\begin{equation}\label{eq9}
\begin{aligned}
&(I_{i}^{eth}, [x_i^{r}, y_i^{r}]), i=1,2,...,m \times n \\
&(I_{i}^{eih}, [\Delta x_i^{r}, \Delta y_i^{r}, \Delta z_i^{r}]), i=1,2,...,m \times n
\end{aligned}
\end{equation}

\subsubsection{Calibration and Automate Label}

For eye-to-hand, the transformation between camera and robot, ${}^cT_r$ and ${}^rT_c$, need to be calibrated for automated label and vision-based localization in robot control. Considering only $x$ and $y$ dimensions, the transform can be formulated as Eqn. \eqref{eq10} and \eqref{eq11}.

\begin{equation}\label{eq10}
[x_i^{c}, y_i^{c}]^T =
\left[
\begin{array}{ccc}
    a_{11} & a_{12} & a_{13} \\
    a_{21} & a_{22} & a_{23} 
\end{array}
\right] [x_i^{r}, y_i^{r}, 1]^T 
\end{equation}

\begin{equation}\label{eq11}
[x_i^{r}, y_i^{r}]^T =
\left[
\begin{array}{ccc}
    b_{11} & b_{12} & b_{13} \\
    b_{21} & b_{22} & b_{23} 
\end{array}
\right] [x_i^{c}, y_i^{c}, 1]^T 
\end{equation}

For eye-in-hand, we mainly focus on the mapping relationship of relative motion between the robot and the task in the robot coordinate system and pixel coordinate system, to realize semi-automatic annotation. Because the image Jacobian can be considered as constant in a limited space, the transform $J$ can be formulated as Eq. \eqref{eq12}. 

\begin{equation}\label{eq12}
\left[
\begin{array}{cccc}
{}^l \Delta x_i^{c}\\
{}^l \Delta y_i^{c}\\
{}^r \Delta x_i^{c}\\
{}^r \Delta y_i^{c}
\end{array}
\right ]
=
\left[
\begin{array}{cccc}
    c_{11} & c_{12} & c_{13} & c_{14} \\
    c_{21} & c_{22} & c_{23} & c_{24} \\
    c_{31} & c_{32} & c_{33} & c_{34} \\
    c_{41} & c_{42} & c_{43} & c_{44} 
\end{array}
\right] 
\left[
\begin{array}{cccc}
 \Delta x_i^{r} \\
 \Delta y_i^{r} \\
 \Delta z_i^{r} \\
 1
\end{array}
\right ]
\end{equation}

Using a small number of coordinates in the pixel frame $L_{eth}^i = [x_i^{c}, y_i^{c}]$ and $L_{eih}^i = [{}^l \Delta x_i^{c}, {}^l \Delta y_i^{c}, {}^r \Delta x_i^{c}, {}^r \Delta y_i^{c}]$ provided by manual annotation and coordinates in the robot frame captured during sampling, the transformation from the robot base frame to the eye-to-hand pixel frame $^cT_{r}$, $^rT_{c}$ and the relative motion relationship between the robot base frame and the eye-to-hand pixel frame can be estimated. The calibrated transform relationship can be used to automate the labeling of the remaining images. In addition, the transform $^rT_{c}$ from the pixel frame to the robot's base frame can be used as hand-eye-task calibration, which involves estimating the assembly pose of the robot's end-effector for localization by the eye-to-hand camera.

\subsubsection{Fine-tuning from Pre-trained Model}
This work used a one-stage real-time object detection approach, YOLO (You Only Look Once), to estimate assembly goal pose and visual attention, which is famous for robustness and fast computation. In addition, the pre-trained model using the ImageNet dataset can be used as initial parameters for training in custom datasets, which require fewer samples. The image and labels are divided into training and test sets for model training and evaluation.
%The structure of one-stage detection is simpler than the fast R-CNN and obtains equivalent accuracy. 

\subsection{Classroom-to-Real Residual Reinforcement Learning}
In this subsection, we discuss a practical Residual Reinforcement Learning for fine manipulation to address the challenges posed by exploration efficiency and safety in semi-structured environments. Classroom-to-real learning trains the residual policy within a simplified structured environment and subsequently transfers it to semi-structured environments. The visual representation and coarse operation provide a base policy and task-relevant features for context generalization to facilitate effective learning and seamless transfer.  
%where the task board is secured by a fixture. To bolster robustness, a curriculum incorporating random errors is injected into the motion guidance, preparing the system to handle real-world uncertainties effectively. 

\begin{figure}[!t]
\centerline{\includegraphics[width=\columnwidth]{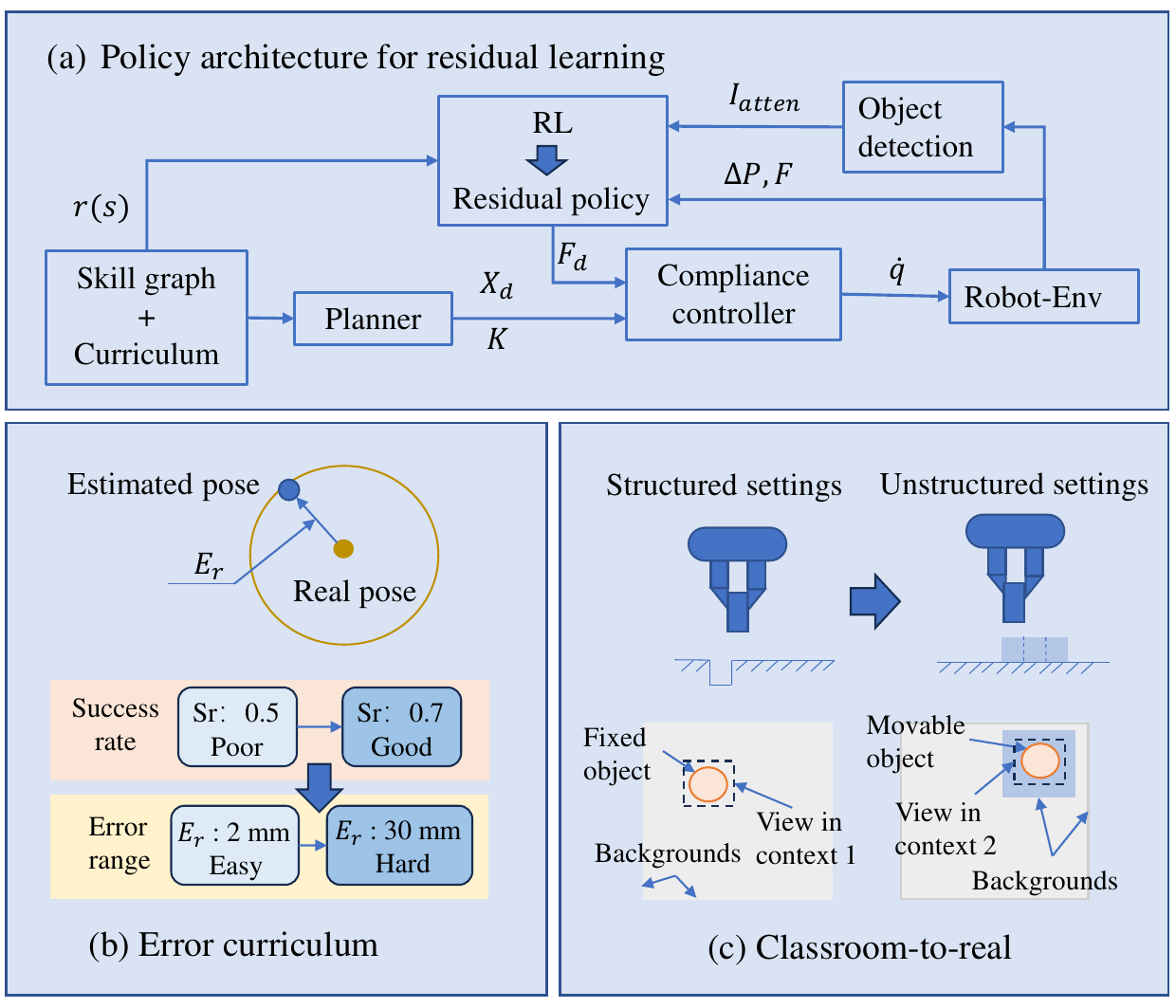}}
\caption{Classroom-to-real learning trains the residual policy within a simplified structured environment and subsequently transfers it to semi-structured environments. (a) shows the cognitive manipulation architecture for residual policy training in a structured environment. (2) introduces the error curriculum for robustness to the uncertainty of coarse operation and improved exploration for efficient learning. (c) shows the residual policy is trained in a structured setting (classroom) and transferred to semi-structured settings (real scene). }
\label{fig5}
\end{figure}

\subsubsection{Curriculum Residual Learning}
\textcolor{blue}{In a structured environment, the key poses of the end effector can be obtained through demonstration. The coarse operation as a base policy is generated and local task detection as attention is loaded for the initialization of the cognitive manipulation.} This work formulated the combination of base and residual sub-policies based on the compliance controller in task space as follows: 
\begin{equation}\label{eq5-4-C-1}
\dot q = f_{cr}(\pi_H, \pi_\theta)
\end{equation}

To increase the robustness of residual policy to the perceptual uncertainty of the pose estimator, a random error is injected into the trajectory. However, the error range may affect the learning efficiency due to the difficulty of exploration and the diversity of samples, as shown in Fig. \ref{fig5}. Therefore, this work automatically controls the task difficulty by increasing or decreasing $\varepsilon$ to the guidance error range of $E_{r0}$ to keep the success rate $s_r$ within the desired interval $[\alpha, \beta]$, as shown in Eqn. \eqref{eq14}.

\begin{equation}\label{eq14}
E_r = E_{r0} + \varepsilon*1_{s_r>\beta} - \varepsilon*1_{s_r<\alpha}
\end{equation}

\subsubsection{Reward Shaping}
This work normalizes the Euclidean distance between the end-effector current $X$ and target pose ${}^BX_E^g$ to create a guide reward $R_{guid}$ that increases as it gets closer to the target pose as in Eqn. \eqref{eq13_1}. The force penalty $R_{forc}$ is defined according to the interaction force $F$ for smooth operation as in Eqn. \eqref{eq13_2}. Additionally, if the transition condition $c_4$ is satisfied, it gets a positive reward of $R_{succ}$ as in Eqn. \eqref{eq13_3}. \textcolor{blue}{Therefore a multi-objective reward function is defined as Eqn. \eqref{eq13}, which uses weights $\lambda_1$, $\lambda_2$, and $\lambda_3$ to balance multiple sub-objectives.}

\begin{equation}\label{eq13_1}
R_{guid} = \left \| diag(W)^{-1} (X - {}^BX_E^g) \right \|
\end{equation}

\begin{equation}\label{eq13_2}
R_{forc} = \left \| diag({F}^{\max })^{-1} F  \right \|
\end{equation}

\begin{equation}\label{eq13_3}
{R_{succ}, d} = \begin{cases}
100,1 &{\text{if}}\ c_4 \\ 
{0,0}&{\text{otherwise.}} 
\end{cases}
\end{equation}

\begin{equation}\label{eq13}
r(s) = \lambda_1 R_{guid} + \lambda_2 R_{forc} + \lambda_1 R_{succ}
\end{equation}

\subsubsection{Soft-Actor-Critic}
A model-free DRL, soft actor-critic, is introduced to achieve a real-time optimal control strategy for intricate fine manipulation. Unlike pure RL, residual learning enhances the performance of the entire policy by optimizing the residual parameterized part of the policy. The state and action of the residual policy, along with the reward of the overall policy, are gathered in multiple recurring episodes and stored in a data replay buffer for off-policy learning. The training was carried out in a structured environment and then transferred to a semi-structured environment. The cognitive manipulation algorithm is depicted in Aglo.\ref{alg1}, where Line 1-2 acquire the coarse operation $\pi ^ {cf}_H$ and $\pi ^ {cr}_H$ with the estimated localization ${}^BX_E^g$, Line 3 -9 obtain the desired pose, $X_d$, stiffness $K$ and desired force/torque $F_d$, Line 6 drives the robot to the assembly bottleneck pose, Line 11-18 perform fine manipulation for accurate assembly tasks.

\begin{algorithm}
    %\textsl{}\setstretch{1.8}
	\renewcommand{\algorithmicrequire}{\textbf{Input:}}
	\renewcommand{\algorithmicensure}{\textbf{Output:}}
    \caption{Cognitive manipulation for robotic assembly} 
	\label{alg1} 
	\begin{algorithmic}[1]
		\REQUIRE $\Delta P_g$, $\Delta P_p$, $X$, $F$, $I_{eth}$, $I_{eih}$
        \ENSURE $X_d$, $F_d$, $K$
		\STATE Estimate localization $P$ with Eqn. (1-4)
		\STATE Planning motion guidance $\pi ^ {cf}_H$ and $\pi ^ {cr}_H$ with Eqn. (6-9)
		\FOR{each time step}
		\STATE $X_d, K \gets \pi ^ {cf}_H$
		\STATE $F_d \gets 0$
        \STATE Apply action $X_d$, $F_d$ and $K$ to robot controller
        \IF {$|X-P_p|\leq E_c$} 
          \STATE Break
        \ENDIF
        \ENDFOR
        \FOR{each time step}
        \STATE Estimate attention $I_{atten}$ with Eqn. (5)
		\STATE $X_d, K \gets \pi ^ {cr}_H$
		\STATE $F_d \gets \pi _ {\theta}$
		\STATE Apply action $K$, $F_d$ and $X_d$ to robot controller
		\IF {$|X-P_g|\leq E_c$} 
          \STATE Break
        \ENDIF
		\ENDFOR
	\end{algorithmic} 
\end{algorithm}

\section{Experiment}
This section delineates the experimental validation of the proposed cognitive manipulation method, specifically designed for robotic assembly tasks within semi-structured environments. 
%The primary objective of these experiments is to rigorously evaluate the proposed semi-supervised visual representation and classroom-to-real fine manipulation techniques to demonstrate the distinct effectiveness of skill graph-based Top-Down learning and cognitive manipulation. 
\textcolor{blue}{Firstly, we introduce the robot hardware and software and establish several baselines to compare the proposed method with existing methodologies. Secondly, comparison and ablation experiments are performed in simulation to validate hand-eye calibration and auto-annotation methods using small amounts of manually annotated data, the effect of embodied data acquisition on the object detection model, and the effect of object detection on the training and performance of fine manipulation in semi-structured environments by providing location and visual attention. Finally, the comprehensive evaluation with two real tasks underscores the practical applications of our approach for robotic assembly within semi-structured environments.} %In comparison to several baselines, the simulation experiments first aim to study semi-supervised visual representation and demonstrate the effectiveness of separate learning for cognition and manipulation. Additionally, we assess the robustness of the system to variable backgrounds through motion and observation guidance.

\subsection{Experiment Setup}
\subsubsection{Hardware and Software}
\textcolor{blue}{The experiments are conducted on a computer equipped with an Nvidia GeForce RTX 2060 GPU and an Intel i7-9700 CPU. The Robot Operating System (ROS) is utilized as the middleware, facilitating seamless communication between the learning algorithms, control modules, and the robotic system. }

%\subsubsection{Evaluation Metrics} 
%\textcolor{blue}{To rigorously evaluate the sample efficiency and context generalization of the proposed method, we employ the following metrics: (1) Standard Deviation (STD): Statistical indicator of the consistency of calibration results. (2) mean Average Precision (mAP@.5:.95): The average precision mean of the model's detection accuracy over a range of different intersection-to-union (IoU) thresholds from 0.5 to 0.95. (3) Cumulative Reward (CR): The total reward accumulated over an episode. (4) Error Range (Er): The deviation from the target position during task execution. (5) Success Rate (SR): The percentage of tasks completed successfully. (6) Completion Time (CT): The average time taken to complete a task.} 

\subsubsection{Partial Model, Data Collection and Reward Design}
\textcolor{blue}{The geometry information for data collection in semi-supervised object detection models and classroom-to-real fine manipulation policy training is obtained by demonstration. Maximum contact force $F^{max}$ is set as 10 N in x, y and z directions and 0.1 N*m in Rx, Ry and Rz directions. The hybrid policy updates the pose and force commands to the controller at a frequency of 5 Hz, while the controller outputs the target joint velocities for the robot at 120 Hz. Each experimental episode is capped at 120 steps, with policy networks undergoing 200 gradient updates per episode. The reward weights $w_1$, $w_2$, and $w_3$ are set to 1, 0.8, and 1, respectively, as determined by preliminary experiments to balance operation speed and smoothness. The curriculum increases or decreases by 0.5 mm to the error range from 2 mm to keep the success rate within the desired interval [0.5, 0.7].}

\subsubsection{Baselines for Comparative Study}
To underscore the advantages of our cognitive manipulation architecture in terms of learning efficiency and context generalization, we compare our method in assembly tasks against the following baselines: 
\begin{enumerate}
\item{Baseline 1 \cite{2020LearningTA}: This baseline directly predicts the desired final poses of slave object for manipulation using only raw RGB-D images and implements a simple open-loop control scheme on a real robot. The training data for pose estimation is synthetically generated, facilitating easy manipulation of geometry and texture. } % Pose-Based Visual Servoing with Object Detection (MB-Yolo-RS) uses a fixed policy that combines trajectory planning and compliance control, and object detection as a pose estimator to guide servoing toward the hole for successful insertion.
\item{Baseline 2 \cite{Zhang2023IntegratingAP, mou2022poseb15}: This approach leverages direct teaching to specify global master object poses and plans robotic motion accordingly. It utilizes hand-mounted cameras and visual classifiers to predict and correct positioning errors, enabling precise attachment of master and slave objects without calibration by search policy.} %Image-Based Visual Servoing with Object Detection (MB-Yolo-RS) uses a fixed policy that combines trajectory planning and compliance control, and object detection as a pose estimator to guide servoing toward the hole for successful insertion. search policy.
\item{Baseline 3 \cite{lee2020guidedb16}: This baseline introduces a perception system with uncertainty estimates to delineate regions where the model-based policy is reliable from those where it may be flawed or undefined, blending the strengths of model-based and learning-based methods. }  %by dividing the manipulation into two stages. First, it uses the policy from MB-Yolo to achieve initial reach and coarse alignment within a defined safe region, and second, it uses a parameterized policy learned through DRL for fine alignment and insertion, thereby addressing inaccuracies in the perception and actuation pipeline of robotics.
\item{Baseline 4 \cite{shi2021proactiveb17}: This work combines a vision-based fixed policy with a contact-based residual parametric policy, enhancing the robustness and efficiency of the RL algorithm.} %uses a hybrid method similar to GUAPO. However, MB-Yolo is extended to the contact-rich region, and a residual force-based parametric policy is used to adapt to reasonable variations during the fine alignment and insertion stages. (vision representation).
\item{Baseline 5 (\cite{Ranjbar2021ResidualFL}): This baseline Employs similar residual learning techniques with Cartesian impedance control, utilizing visual inputs for larger error adjustments during contact-rich manipulation. }  %uses a multimodal residual policy to account for servo error and address friction during the insertion process. Unlike the proposed approach, observation guidance is not applied
\end{enumerate}

\subsection{Simulation Experiment for Comparative and Ablation Study}
Our proposed approach seeks to enhance the sampling efficiency and reduce the engineering effort required for policy reconfiguration in contact-rich tasks within semi-structured environments. This is accomplished by integrating semi-supervised learning of object detection with classroom-to-real residual reinforcement learning of fine manipulation. \textcolor{blue}{To facilitate a comparative and ablation study, we have developed a simulation environment based on Gazebo, which allows for dynamic loading and deletion of objects at any position within the defined space, thereby constructing a robot assembly task in a semi-structured setting. The application of our proposed cognitive manipulation framework to a new assembly task is structured into four distinct stages: Embodied hand-eye-task calibration and semi-automatic annotation, supervised fine-tuning of the object detection models, residual reinforcement learning of fine manipulation, and application of the integrated strategy to a semi-structured environment. Each stage is progressively compared with established baselines to evaluate the advantages of the proposed methodologies.}
%The two components are trained separately to compare their performance and assess the strengths and weaknesses of each approach. We first train the two object detection models with a collected custom dataset by fine-tuning the pre-trained model. Then, using the motion and observation guidance provided by the trained cognition model, the training process of the cognition-guided manipulation policy is compared with several existing RL-based methods to demonstrate its advantages.

\subsubsection{Embodied Hand-Eye-Task Calibration and Semi-Automatic Annotation}
\textcolor{blue}{The dataset is curated based on prior knowledge of uncertainty to ensure sample diversity. The initial stage aims to minimize the costs associated with manual labeling and hand-eye calibration while generating high-quality labeling data and accurately estimating the target assembly pose of the end-effector. This stage encompasses four critical processes: data acquisition, manual labeling, hand-eye-task relationship fitting, and semi-automatic labeling. We explore the dependency on the proportion of manually annotated samples and its advantages over purely manual annotation. The standard deviation of the hand-eye relationship fitting serves as a quantitative index for evaluating the calibration and annotation.}

\begin{table}
\caption{Evaluation of the proposed embodied calibration and annotation by standard deviation.}
\label{table}
\centering
\setlength{\tabcolsep}{3pt}
\begin{tabular}{cccccc}
\hline
\multirow{2}{*}{Number} & \multicolumn{2}{c}{${}^cT_r$ STD} & \multicolumn{2}{c}{$J$ STD} & \multirow{2}{*}{${}^rT_c$ STD} \\
& Manual & All & Manual & All & \\
\hline
8 & 0.0039 & 0.0007  & 0.0062 & 0.0010 & 0.0020  \\	
19 & 0.0043 & 0.0010  & 0.0064 & 0.0014 & 0.0025  \\
38 & 0.0041 & 0.0013  & 0.0068 & 0.0021 & 0.0023   \\
\hline
\end{tabular}
\label{tab1}
\end{table}

\begin{figure}[!t]
\centerline{\includegraphics[width=\columnwidth]{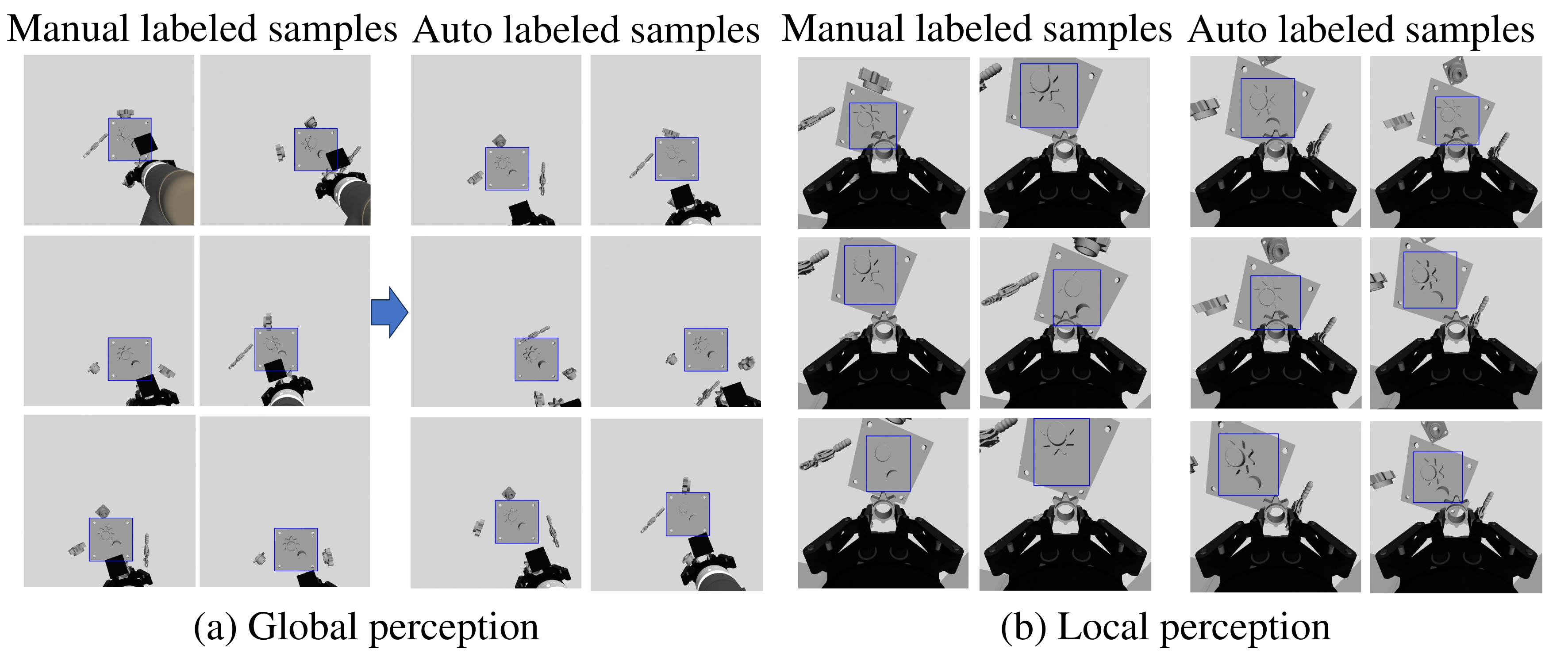}}
\caption{The samples of manual and auto annotation. }
\label{fig6}
\end{figure}

\textcolor{blue}{The number of manually labeled samples and the corresponding accuracy of ${}^cT_r$, $J$, and ${}^rT_c$ represented by standard deviation, as well as the standard deviation of ${}^cT_r$ and $J$ on all data sets are shown in Table \ref{tab1}. The results demonstrate that a small number of manually labeled samples can effectively establish hand-eye calibration relations and facilitate semi-automatic labeling of the remaining images, reducing the standard deviation of data annotation. Examples of both manually and automatically annotated data are illustrated in Fig. \ref{fig6}. } The semi-automatic annotation reduces the cost of manual annotation by 97.4\%.

\subsubsection{Supervised Fine-tuning for Object Detection} 
\textcolor{blue}{In this phase, We examine the impact of sample diversity on model performance by designing various sampling schemes and comparing models trained on different dataset sizes. We mainly consider two factors in the sampling process: pick-and-place position and robot posture, which together affect sample size and diversity, including 10 (5*2), 30 (10*3), 60 (14*5), 120 (15*8), 180 (18*10), 384 (24*16). 
In addition, the combination of 10 (5*2) after data enhancement to 300 is used as a baseline to facilitate the comparison between embodied data acquisition and traditional data enhancement methods based on graph transformation. 20$\%$ of the embodied data enhanced 385 data is taken as the verification set to represent the complex state in the assembly operation process. Precision, recall, mAP@.5, and mAP@.5:.95 are employed to assess the influence of sample quantity and data enhancement methods on the performance of the object detection models. }

%\begin{figure}[!t]
%\centerline{\includegraphics[width=\columnwidth]{Figures/Fig. 4. Hand-designed policy based on the workspace and task information.png}}
%\caption{The sampling points for planning motion to collect diverse samples. Sampling points are designed based on the workspace and task information to plan motion. Image and position information are recorded for calibration and model training.}
%\label{fig4}
%\end{figure}

\begin{figure}[!t]
\centerline{\includegraphics[width=\columnwidth]{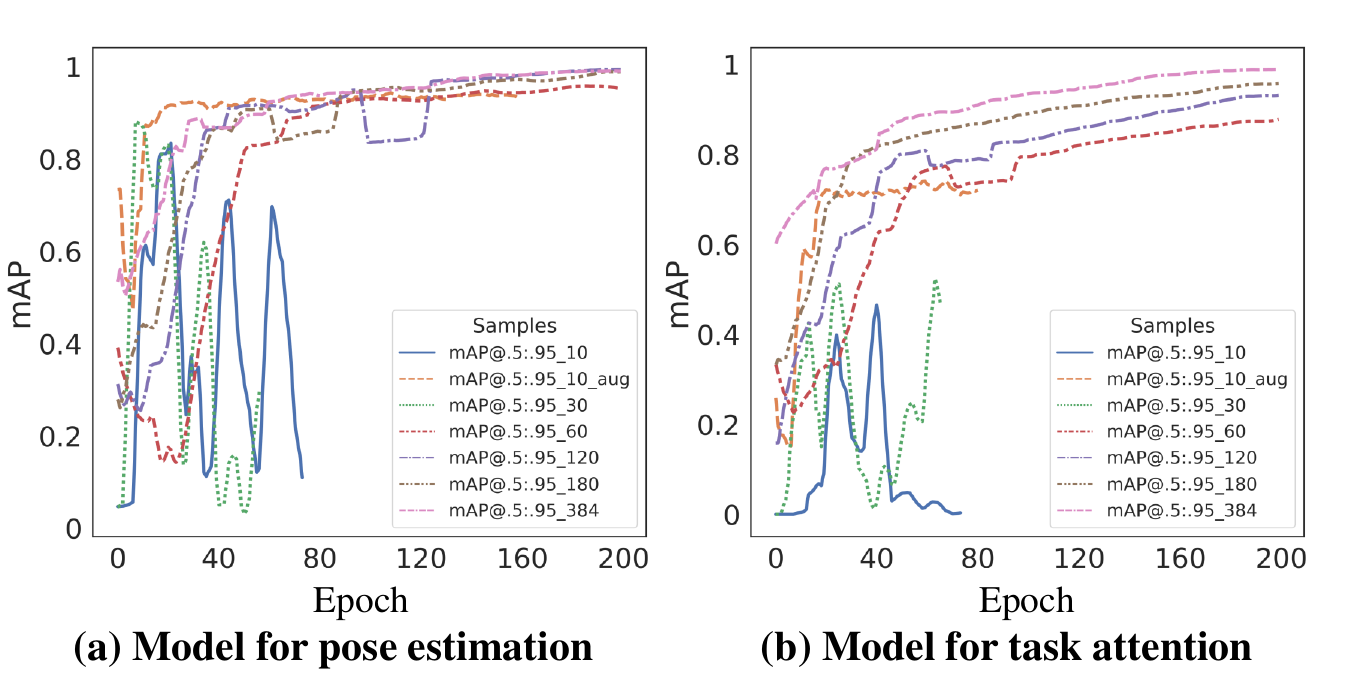}}
\caption{Learning process of YOLO models by fine-tuning with custom datasets. We load the YOLO v8s and fine-tune it for 200 epochs. mAP@.5:.95 is used to evaluate the training process.}
\label{fig6}
\end{figure}

\begin{table}[!t]
\centering
\tabcolsep=0.12cm
\footnotesize
\caption{The performance of learned policies}
\label{tab:table2}
\begin{tabular}{ccccccc}
\hline
Models & Datasets & Precision & Recall & mAP@.5 & mAP@.5:.95  \\
\hline
\multirow{7}{*}{{Pose estimation}} & 10 & 0.233 & 0.448 & 0.156 & 0.109 \\
& 10-aug & 0.999 & 1.0 & 0.995 & 0.935 \\
& 30 & 0.750 & 0.741 & 0.774 & 0.374 \\
& 60 & 0.999 & 1.0 & 0.995 & 0.950 \\
& 120 & 0.999 & 1.0 & 0.995 & 0.994 \\
& 180 & 0.999 & 1.0 & 0.995 & 0.993 \\
& 384 & 0.999 & 1.0 & 0.995 & 0.990 \\

\hline
\multirow{7}{*}{Task attention} & 10 & 0.022 & 0.051 & 0.012 & 0.003 \\
& 10-aug & 0.999 & 1.0 & 0.995 & 0.714 \\
& 30 & 0.686 & 0.980 & 0.815 & 0.539 \\
& 60 & 0.999 & 1.0 & 0.995 & 0.879 \\
& 120 & 0.999 & 1.0 & 0.995 & 0.936 \\
& 180 & 0.998 & 1.0 & 0.995 & 0.964 \\
& 384 & 1.0 & 1.0 & 0.995 & 0.989 \\
\hline
\end{tabular}
\end{table}

\begin{figure}[!t]
\centerline{\includegraphics[width=\columnwidth]{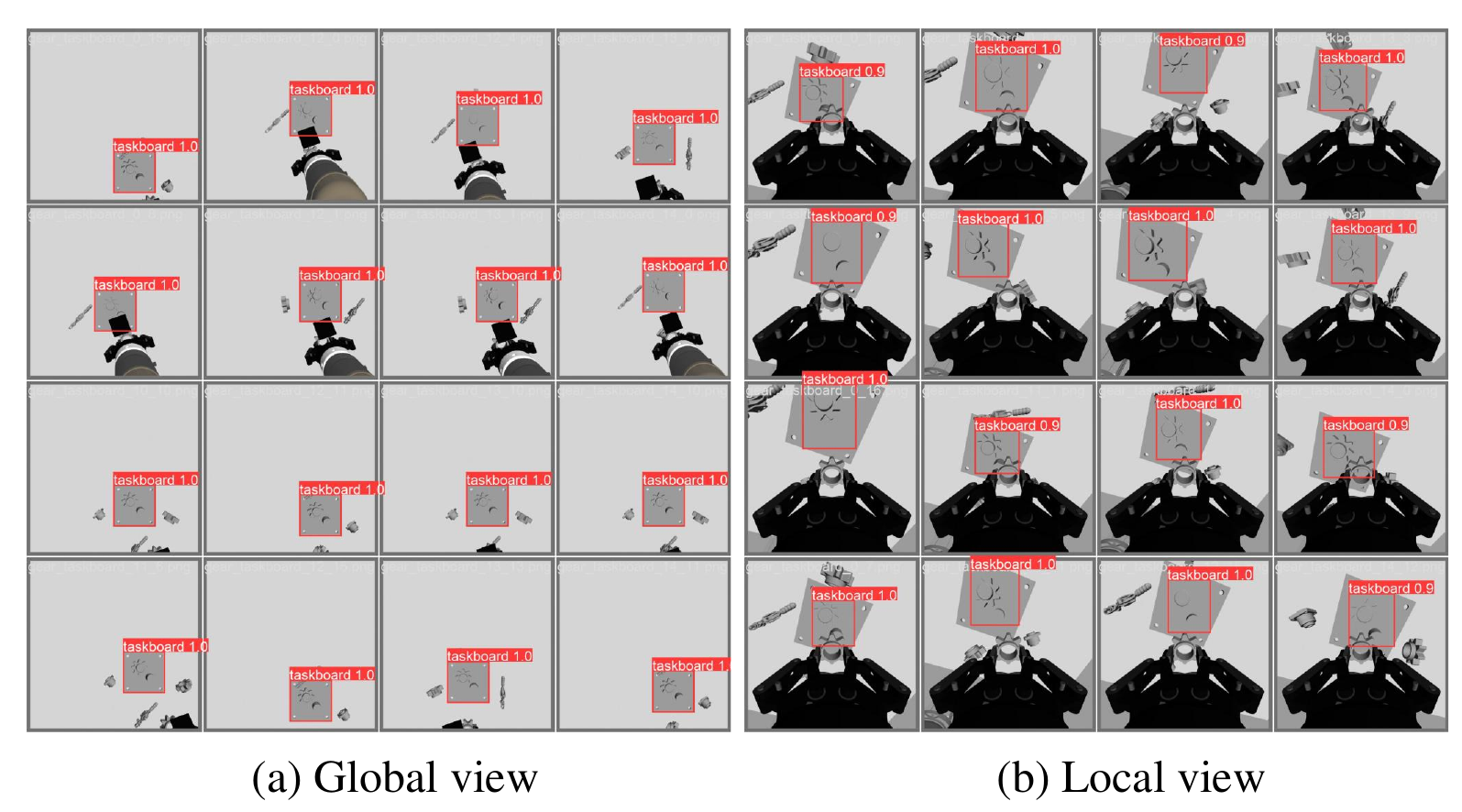}}
\caption{Results of object detection by YOLO models. }
\label{fig6}
\end{figure}

\textcolor{blue}{The results indicate that data acquisition based on prior knowledge significantly enhances the environmental perception capabilities of the detection model.} The mAP@.5:.95 values for the two object detection models during the training process are displayed in Fig. \ref{fig6}, highlighting the importance of sample diversity and the limitations of data enhancement based solely on graphical transformations. \textcolor{blue}{Contrasting the effect of the number of samples on the model performance, too few samples, less than 60, will cause the training process to fail to converge. With the increase in the number of samples, the trained model can obtain higher precision, recall, mAP@.5, and mAP@.5:.95 values. This is due to improved sample diversity by more background changes and robot-task relative posture.} In particular, the performance of local task attention models is more sensitive to sample diversity due to unavoidable occlusion during contact-rich operations. Contrasting methods of sample generation, although data enhancement based on graph transformation can increase the number of samples to avoid overfitting, it lacks diversity. The position estimation model achieved acceptable results, while the task attention model performed poorly. The embodied data collection increases the mAP@.5:.95 by 5.5$\%$ in global perception and 27.5$\%$ in local perception compared to existing data augmentation methods.

%The {mAP-0.95} reaches 0.995 after 83 iterations on the custom dataset, where an Intersection over Union (IoU) greater than 95 guarantees a positioning error of less than 10 mm. But the {mAP-0.95} only converges to 0.945 because of the limited view and the automatic annotation using approximate relations. Both the global localization and local attention models can be trained within 1 hour, including 45 minutes for sampling and 15 minutes for training two models. The trained model can guide the manipulation. %(mAP in different number samples and data aug. visualize the detection result)

\subsubsection{Residual Reinforcement Learning of Fine Manipulation}
\textcolor{blue}{This stage involves training a residual policy, supported by a hand-designed base policy and task-focused view, on a fixed master object setting.%, as displayed in Fig. \ref{fig6}.
Physical contact states are identified using LSTM networks to encode time-series data from touch and proprioception sensors, combined with visual feedback processed through a CNN for serious position error. An MLP then integrates the low-dimensional latent features from LSTM and CNN to generate residual actions. We contrast this procedure with three baselines 3, 4, and 5 to highlight the advantages of knowledge-informed learning.}

%\begin{figure}[!t]
%\centerline{\includegraphics[width=0.8\columnwidth]{Figures/Fig-5-B-3-Robot assembling the object in semi-structured environments.pdf}}
%\caption{Gear assembly task in a semi-structured environment. }
%\label{fig6}
%\end{figure}

\begin{figure}[!t]
\centerline{\includegraphics[width=\columnwidth]{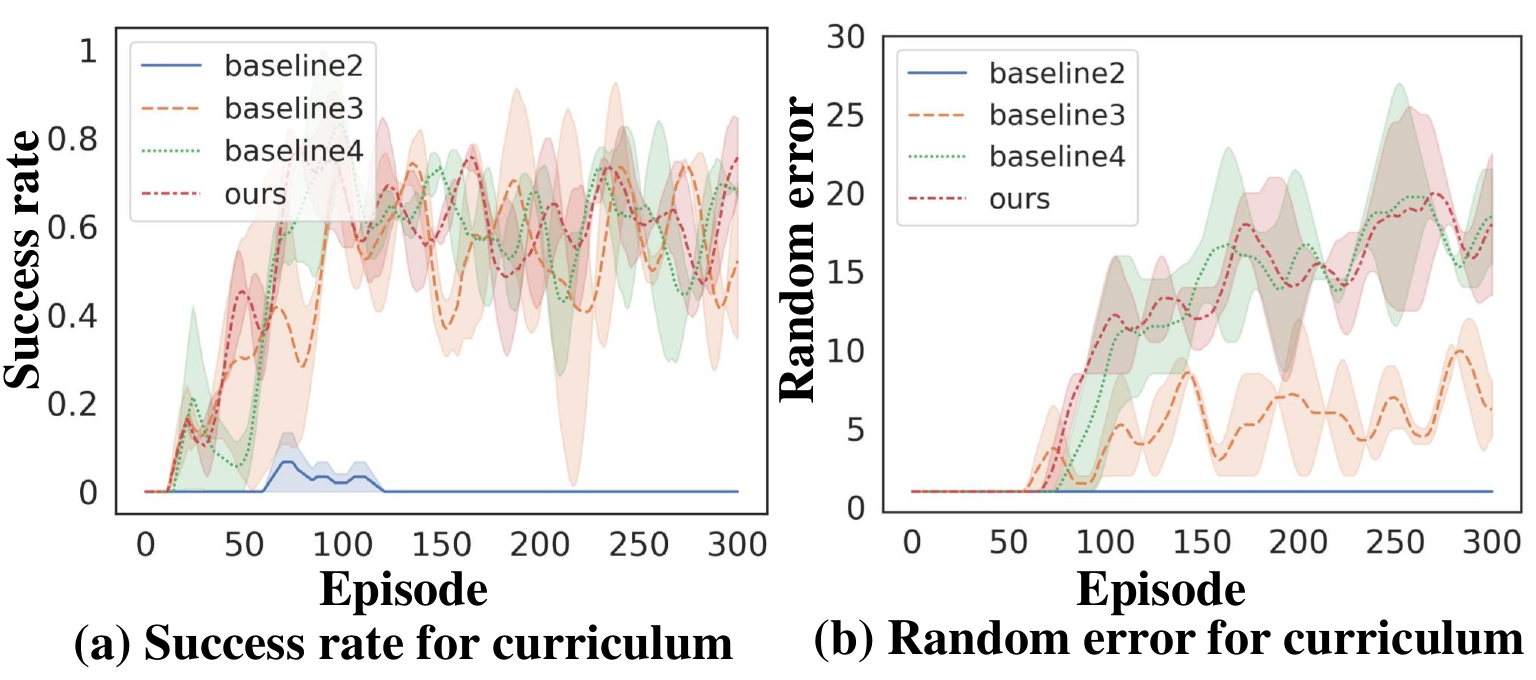}}
\caption{Learning process of residual policy through curriculum residual reinforcement learning for 300 episodes. (a) shows the accumulated reward over an episode. (b) illustrates the error range adjusted by an adaptive curriculum based on the success rate.}
\label{fig7}
\end{figure}

Results suggest that our approach facilitates more efficient and effective learning by concentrating on task-relevant details and addressing intricate contact dynamics and positional uncertainties. The success rate and error range throughout the training of the manipulation policy are presented in Fig. \ref{fig7}. In baseline 3, without the base policy, pure RL struggles in precise insertion tasks due to a local optimum created by penalizing contact forces. Combining the base policy with RL-based residual policy in baseline 4 can result in success, but the curriculum only achieves an 8 mm error in force-based residual policy training due to limited observations of contact force and vision-based pose estimation, creating a Partially Observable Markov Decision Processes (POMDP). In baseline 5, utilizing raw visual information can improve observations, with the curriculum achieving around 20 mm. The ROI-based attention in our approach, allowing the policy to rely on limited features and introducing perturbations due to restricted detection accuracy, marginally influences efficiency. 
%It took 300 episodes to learn a robust residual policy with a fixed policy of 20 mm error, which took 4 hours.

\subsubsection{Cognitive Manipulation in Semi-structured Environment}
After individual training phases, the integrated policy is evaluated for its effectiveness in terms of success rate and completion time during assembly tasks in a semi-structured environment. The manipulator grasps the slave object, a gear, while the master object, a task board, is randomly positioned within a confined workspace of 350*350 mm. We carry out 16 trials to compare our method with other baselines, excluding the non-convergent baseline 3.
%with obstacles by randomly placing the task board in the limited workspace while the gear was placed on a fixture with little uncertainty. The robot was required to grasp the gear, move to the task board, properly align it, and insert it. Our work focused primarily on the alignment and insertion steps in this semi-structured environment. We compared our proposed approach to other task completion baselines, all of which were initiated from a manually designed grasp or re-grasp policy at a specific location outside the workspace. Specifically, we compared our approach to several baselines, excluding the non-convergent baseline 2.

\begin{figure}[!t]
\centerline{\includegraphics[width=0.6\columnwidth]{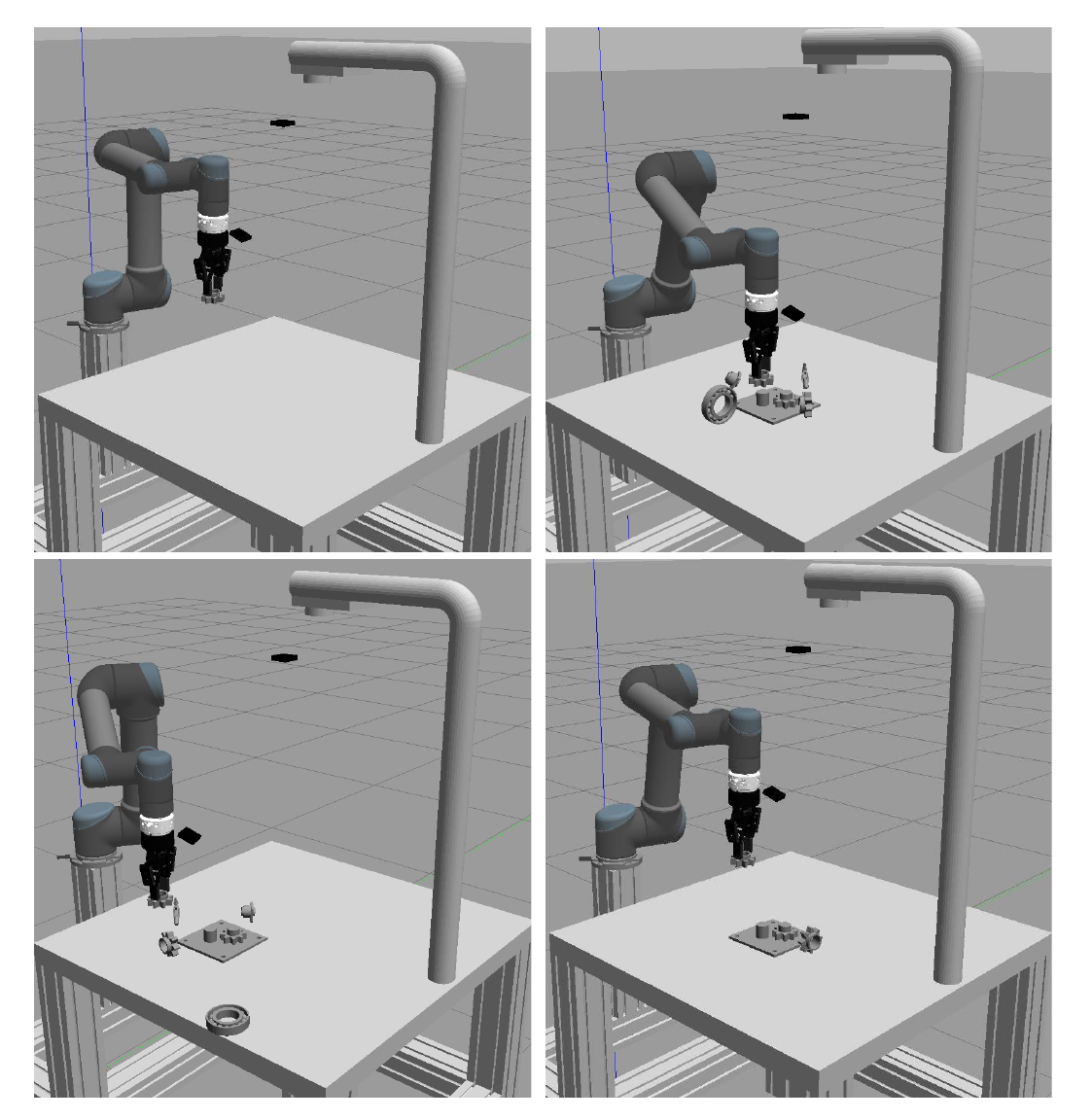}}
\caption{Gear assembly task in a semi-structured environment. }
\label{fig6}
\end{figure}

\begin{table}
\caption{Evaluation of the proposed approach in semi-structured environments by success rate and completion time.}
\label{table}
\centering
\setlength{\tabcolsep}{3pt}
\begin{tabular}{ccc}
\hline
Methods & Success rate & Costed steps \\
\hline
Baseline 1 & 0.125 & $64.03 \pm 18.45$ \\
Baseline 2 & 0.313 & $104.3 \pm 23.63$ \\
Baseline 4 & 0.87 & $85.3 \pm 19.11$ \\
Baseline 5 & 0.67 & $73.4 \pm 32.86$ \\
Ours & 1.0 & $54.4 \pm 17.17$ \\
\hline
\end{tabular}
\label{tab1}
\end{table}

The results indicate that our approach exhibits superior performance in handling challenging assembly tasks, achieving a perfect success rate and significantly reducing completion steps, as outlined in Table \ref{tab1}. Baseline 1 can only complete 12.5 $\%$ of trials with an average of 64 steps, while baseline 2 can only complete 40 $\%$ of trials with an average of 104 steps, nearing the maximum time step. The object detection, single-camera usage, and the simple model-based method fall short of meeting the task's accuracy and the environment's uncertainty requirements. Although random residual actions can help to compensate for the perception errors, the semi-structured environment poses additional challenges due to movable objects. The contact force generated during the random search can displace the task board, leading to larger errors or even causing the gear to slip off the peg. Baseline 4 outperforms baseline 2 in both success rate and cost steps because force-based agents can enhance the search policy by regulating the contact force and position reference based on the estimated contact state derived from interaction forces. Although baseline 5 is more robust than baseline 4 in training, it only performs 13.95 $\%$ better in costed steps and even worse in success rate. The raw visual information enables the residual policy to compensate for larger errors in training, but its performance may degrade in different locations with different backgrounds. In comparison, the proposed method utilizes visual attention to assist the agent in focusing on the task, resulting in a success rate of 1 with an average of 54.4 costed steps. In conclusion, the success rate is increased by 13$\%$ and the number of steps is reduced by 15.4$\%$ compared to competing methods.

\subsection{Comprehensive Evaluation on Real Tasks}
\begin{figure*}[!t]
\centerline{\includegraphics[width=1.6\columnwidth]{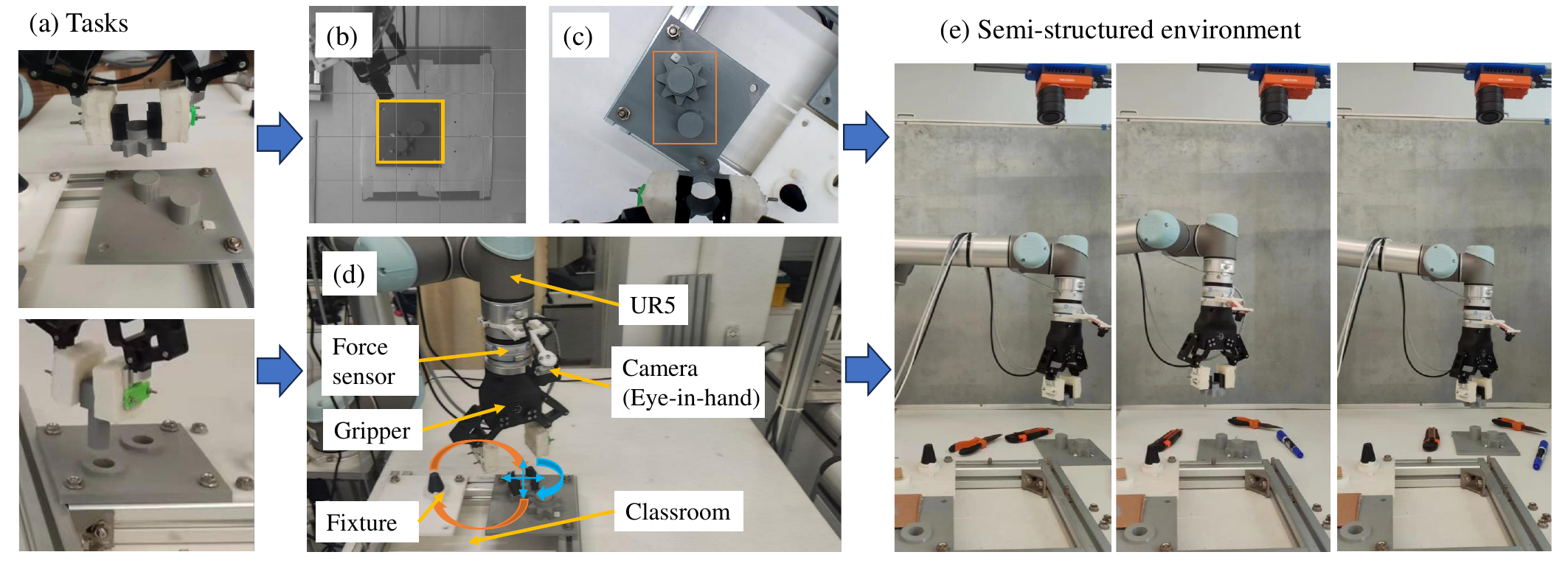}}
\caption{The robotic assembly tasks in a semi-structured environment to evaluate the proposed cognitive manipulation. (a) illustrates the gear-insertion and peg-in-hole assembly tasks examined in this study. (b), (c), and (d) show a structured classroom for robot learning with points for data collection for training the object detection model and fixtures for the RL agent for fine manipulation. (e) displays a semi-structured deployment environment for simulating a flexible manufacturing scenario, with a task board randomly placed in a pre-defined workspace.}
\label{Fig-5-c-1}
\end{figure*}

The primary objective of this research is to develop and validate a cognitive manipulation framework suitable for robot learning in real-world robotic applications. To assess the effectiveness of our proposed architecture, we conducted experiments using a UR5 robot to execute two precision assembly tasks: peg-in-hole and gear-insertion. These tasks, depicted in Fig. \ref{Fig-5-c-1} (a), are designed to test the robot's ability to handle complex manipulations in real settings. The robot was programmed to perform tasks based on geometric information derived from a teaching phase, which was used to construct a skill graph that encapsulates common assembly knowledge. Critical points including grasp and bottleneck pose were identified in semi-structured environments to facilitate hand-eye-task calibration and semi-supervised fine-tuning of object detection, as shown in Fig. \ref{Fig-5-c-1} (b) and (c). In structured environments, critical points including grasp and assembly goal pose guided the learning of contact-rich fine manipulation, as shown in Fig. \ref{Fig-5-c-1} (d). An Object-Embodiment-Centric (OEC) task representation, incorporating home, grasp, bottleneck, and assembly goal points, was employed to reconstruct the basic operational strategy. This strategy was integrated with visual and fine manipulation models to accomplish assembly tasks within a confined area of 500x500 mm, as shown in Fig. \ref{Fig-5-c-1} (e). 

\textcolor{blue}{The training process was optimized based on insights from simulation experiments, focusing on minimizing training costs while maximizing operational efficiency. For object detection, 5 points were gathered from the workspace to improve environmental robustness. We captured five images from a global perspective at each point for pose estimation and an additional 18 images per point from a local view to enhance task-specific attention. This embodied data collection strategy ensured diversity and data enhancement is further used to enhance the robustness against robot pose variations in manipulation. The fine manipulation training extended over 150 episodes, deemed adequate for achieving resilience against uncertainties in the base policy arising from pose estimation errors and unknown contact dynamics. We evaluated the success rate and completion time of the assembly tasks, using these metrics to benchmark the performance of our proposed architecture against two other baselines.}

\begin{figure}[t!]
    \centering
    \includegraphics[width=0.98\columnwidth]{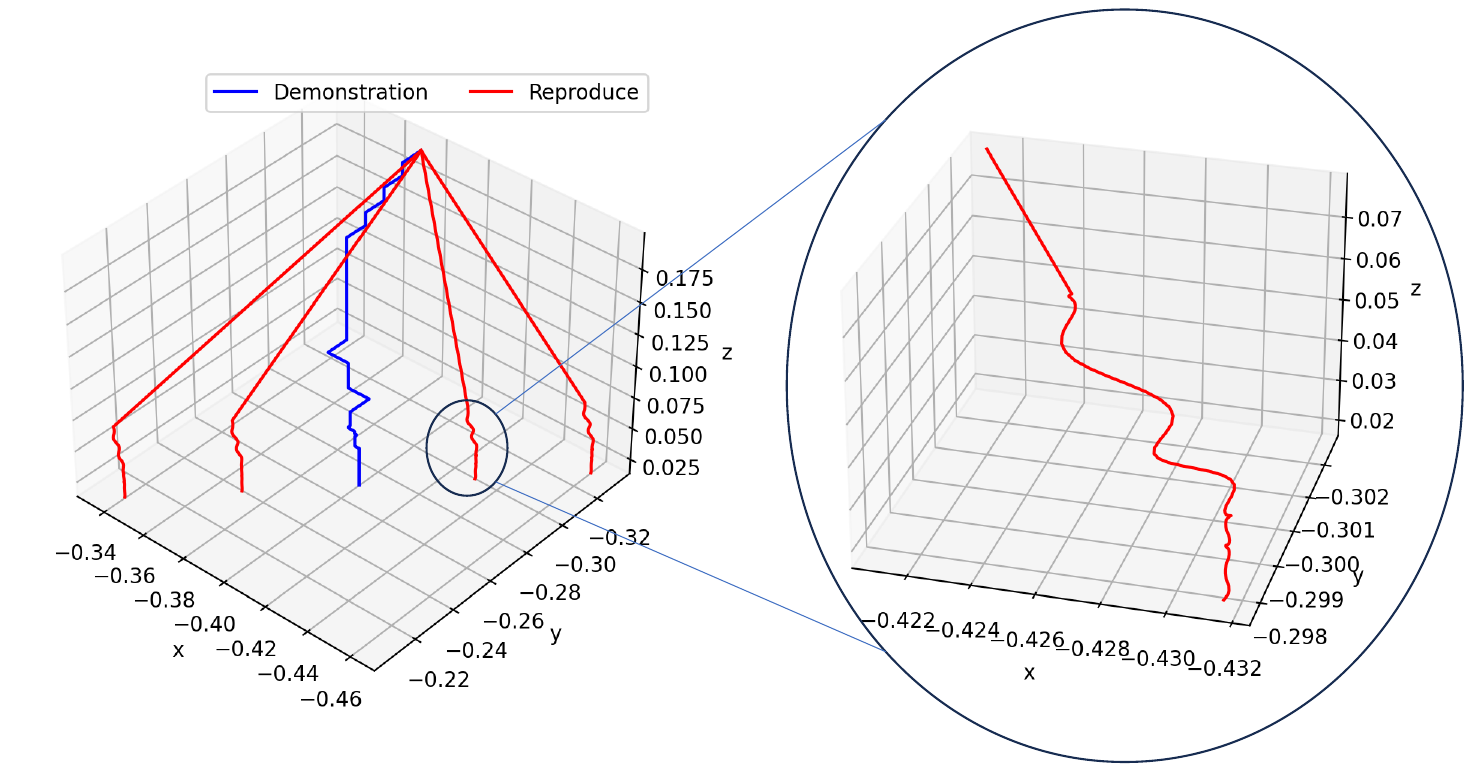}
    \vspace{0.05in}
    \caption{\textbf{Teach and reproduce acquired assembly skills in a semi-structured environment.} \textcolor{blue}{(a) displays three dimensions of trajectories that were demonstrated and reproduced, illustrating the adaptability to variations in the target position. (b) shows the trajectory in contact-rich regions, emphasizing the significance of residual policy.} }
    \label{fig: IL}
\end{figure}

\begin{table}[!t]
\centering
\tabcolsep=0.12cm
\footnotesize
\caption{The performance of learned policies}
\label{tab:table2}
\begin{tabular}{ccccccc}
\hline
Task & Methods & Success Rate & Completion Time  \\
\hline
\multirow{3}{*}{Peg-in-hole} & Baselines1 & 0.18 & $8.13\pm4.96$ \\
& Baselines2 & 0.437 & $18.06\pm6.65$ \\
& Ours & 0.937 & $\ 6.11\pm1.32$ \\

\hline
\multirow{3}{*}{Gear-insertion} & Baselines1 & 0.06 & $9.2\pm3.65$ \\
& Baselines2 & 0.313 & $19.10\pm5.75$ \\
& Ours & 0.875 & $\ 7.04\pm1.42$ \\
\hline
\end{tabular}
\end{table}

The results, detailed in Table \ref{tab:table2}, indicate that our approach significantly outperformed the baselines in both tasks. 
%In the peg-in-hole task, our method increases the success rate by 60$\%$ and reduces the completion time by 50$\%$ with respect to baseline1, and increases the success rate by 60$\%$ and reduces the completion time by 50$\%$ with respect to baseline2. In the gear-insertion task, the success rate is increased by 60$\%$ and the completion time is reduced by 50$\%$ with respect to the baseline1, and the success rate is increased by 60$\%$ and the completion time is reduced by 50$\%$ with respect to the baseline2. In summary, the success rate is increased by 60$\%$ and the number of steps is reduced by 50$\%$ compared to widely used multi-stage methods.
Baselines 1 and 2 faced challenges with the tasks, primarily due to inaccuracies in pose estimation and control, as well as their inability to effectively navigate the semi-structured environment. In addition, traditional search-based methods proved ineffective due to the mobility of the master object, often causing the robot to get stuck on the object's surface. Performance relies on the expert's experience and parameter tuning. It took approximately 8 hours and multiple attempts to gather samples and fine-tune policy and controller parameters for new tasks, whereas our method only required 2.58 hours and minimal human intervention. Both global localization and local attention models can be trained within 0.58 hours, including 20 minutes for sampling and 15 minutes for training two models. Learning a robust residual policy for robustness to 15 mm error took 150 episodes, consuming 2 hours. The experimental results underscore the superiority of our cognitive manipulation framework in the practical applicability in complex, real-world environments, improving learning efficiency and engineering effort while showcasing the precision and efficiency of manipulation for robotic assembly tasks in real-world scenarios.

\section{Discussion}
\textcolor{blue}{The architecture presented in this study leverages a skill graph that merges the generalization capabilities of a pre-trained object detection model with the optimization ability of reinforcement learning to facilitate efficient learning with minimal reliance on extensive human knowledge and interaction data.} \textcolor{blue}{Separate training of different components has proven more practical for real robots in precise assembly tasks compared to end-to-end training methods \cite{Yasutomi2023VisualSA}.} This method mirrors human learning processes, where theoretical knowledge is acquired first and then practiced in controlled settings before tackling complex real-world tasks. 
The skill graph not only enhances the efficiency of the object detection learning process by providing explicit prior knowledge for sample collection and annotation but also enables the system to estimate the assembly target pose similar to \cite{2020LearningTA}. Motion planning guided by the skill graph enables diverse data collection from various perspectives and locations, reducing the need for manual labeling through low-cost calibration and automated labeling processes. The large pre-trained model benefits from this setup by allowing effective generalization through fine-tuning with implicit prior knowledge.
Furthermore, the learned visual model excels in providing interpretable spatial location and task correlation information, surpassing structured visual representations \cite{Zhang_2023} and uncertainty-aware pose estimation \cite{lee2020guidedb16}. This information is essential for guiding and constraining the exploration process in reinforcement learning, enabling the system to efficiently learn about contact dynamics and pose uncertainty. the residual policy, guided by the base policy and focused multimodal observation, is optimized through a multi-objective reward system, enhancing the capability to tackle complex tasks and generalize across various contexts without the need for fixtures. Therefore, the separation and guidance for learning tasks using prior knowledge significantly enhances learning efficiency in controlled environments. 
% is more sample-efficient for data collection and manual labeling in new tasks. The concepts of observational attention and , which are directly defined according to geometric information, can provide data acquisition strategies and labels that allow a passive approach to efficient vision model learning. 

\textcolor{blue}{Compared to the existing combination of model-based and learning-based approaches \cite{lee2020guidedb16, shi2021proactiveb17, Zhao2023ALT}, our cognitive manipulation method excels in semi-structured environments. It mimics the human approach of transitioning from global to local perception and from coarse to fine manipulation.}  In contact-free areas, the object detection estimates the position uncertainty of the main object problem caused by fewer constraints. The skill graph enables global perception beyond the workspace to avoid occlusion and directs coarse operations with rich geometric information, facilitating flexible and safe robot movement. In contact-rich areas, object detection provides visual information about the task's attention and resolves the variable background interference caused by other dynamic objects. The residual strategy integrates task-focused visual and tactile information to solve the pose estimation error and the complex contact dynamics.
 
\textcolor{blue}{The partial models within our method are utilized to address diverse configurations in semi-structured environments. While this study primarily focuses on knowledge-driven robot learning and experiments validate the impact of such learning on efficiency and strategy robustness, this approach can be adapted to different environments by acquiring geometric information through teaching and adjusting temporal logic and transition conditions within the partial model.}

%\textcolor{blue}{Compare to joint training \cite{Yasutomi2023VisualSA}, unsupervised visual representation learning \cite{Zhang_2023}, and supervised representation learning \cite{lee2020guidedb16}, knowledge-guided separate learning is more suitable for real robots and precise assembly tasks. Semi-structured visual representation learning is more sample-efficient for data collection and manual labeling in new tasks. The learned visual representation can provide master object spatial location and task correlation information for operational policies, which is more interpretable than latent or image space features. with the structured visual representation, the architecture is more well-structured and flexible, compared to the existing combining model-based and learning-based approaches \cite{lee2020guidedb16, Zhao2023ALT}. On one hand, the base policy and task-related visual information enable efficient generalizable residual policy for fine manipulation. On the other hand, the master object spatial location enables the scale of the base policy to different locations for assembly tasks in a semi-structured environment.  Therefore, the human-inspired learning and manipulation process allows the efficient learned vision model and model-based base policy to provide more guidance for manipulation policy learning. }

\section{Conclusion}
This study introduces a novel cognitive manipulation framework for robotic assembly tasks in semi-structured environments. The framework employs a skill graph that integrates object detection, coarse operation planning, and fine operation execution. The training process, guided by skill maps and coarse-operation planning in controlled environments, involves semi-supervised learning for object detection and residual reinforcement learning of multimodal fine-operation strategies. The cognitive manipulation models are subsequently transferred to a semi-structured environment, where object detection and coarse operation, enhanced by skill graph, handle the uncertainty of the environment and provide guidance for residual policy to address pose estimation and contact dynamics uncertainty.
%For the training process, guided by skill maps and coarse-operation planning in a controlled environment, we first perform semi-supervised learning for target detection, and then efficiently learn multimodal fine-operation strategies guided by coarse-operation and task attention. The cognition and manipulation models are trained separately with a hand-designed policy in a structured environment and transferred to the semi-structured environment. 
Experimental results from simulation demonstrate that our cognitive manipulation facilitates reducing manual annotation costs by 97.4$\%$ and enables learning assembly tasks involving a 20 mm error and a 0.1 mm gap within 300 episodes, showing significant progress in a semi-structured environment—an area where existing methods struggle—with a 13$\%$ increase in success rate and a 15.4$\%$ reduction in completion time. The practicality of the method was further confirmed in real experiments.
%The results of the experiment on a gear assembly task show that the cognition model and the guided residual model can be learned efficiently. And the cognition-guided manipulation method can perform efficient operations and is robust to robot system errors and variable environments. Semi-supervised learning method can achieve low-cost training of visual perception model, which can ensure the diversity of samples and increase the mAP by 50$\%$ compared with existing data enhancement methods, and reduce the cost of manual annotation by 97.4$\%$ compared with manual annotation. Simultaneous learning of error and contact force strategies for an assembly task with 20 mm error and 0. 1mm gap can be migrated to a semi-structured environment, which is very difficult for existing methods, with a 30$\%$ increase in the success rate and a 20$\%$ reduction in the assembly time compared to model-based methods.

% 我们定义了一个技能图去整合目标检测, 粗操作规划和精操作. 并且在技能图和粗操作规划的引导下在可控的环境中先进行目标检测半监督学习，然后在粗操作和注意力的引导下高效的学习多模态的精操作策略。半监督学习方法可以实现低成本下的视觉感知模型的训练,相对比现有数据增强方法可保证样本的多样性,将mAP提高50%, 相对于人工标注可减少人工标注成本97.4%。在20mm误差和0.1mm间隙的装配任务同时学习误差和接触力策略，并可以迁移到半结构化环境中, 这对于现有的方法是非常困难的,相对于model-based方法,成功率提高了30%,装配时间减少了20%.

Despite these advancements, the learning efficiency and generalization capabilities in semi-structured environments have been substantially enhanced, yet challenges persist. Our method effectively utilizes prior knowledge to streamline the learning process for contact manipulation, especially simplifying the reinforcement learning challenges associated with uncertainties in pose estimation and contact dynamics. However, there is potential to further improve learning efficiency. Future work will focus on advancing learning efficiency through the application of offline enhancement methods, including sim-to-real transfer or meta-learning, to streamline the residual reinforcement learning process. The efficient learning method using prior knowledge increases the possibility of contact-rich manipulation multitasking or meta-learning in real robots. In addition, our approach, which utilizes object detection and skill graphs, aims to mitigate uncertainties in semi-structured environments, but further generalization to diverse environments and tasks remains a goal. Future research could explore more sophisticated 3D or 6D pose estimation techniques, develop more precise quality estimation and monitoring methods based on visuo-tactile fusion, and incorporate large language models (LLMs) for common sense reasoning. Considering complex state and exception handling methods, such as addressing failure and success scenarios, could reduce assumptions about semi-structured environments and enhance the quality and reliability of robot operations.

%Although learning efficiency has been improved through separate learning processes, the learning process for contact-rich manipulation, which currently takes 4 hours, still needs further improvement. In addition, the use of a distance-based success criterion poses challenges for assessing assembly quality without precise localization. In future work, we aim to improve the efficiency of reconfiguration by exploring methods such as object detection with data augmentation and using DRL with pre-trained neural networks transferred from other tasks or simulation environments. We plan to investigate quality estimation methods that can replace the current distance-based criterion. This will involve considering factors such as the error in the estimated target pose and determining the appropriate range of criterion values that will determine the quality of completion. 

%use a bibliography generated by BibTeX
\bibliography{Cognitive-manipulation}{}
\bibliographystyle{IEEEtran}

\vspace{2pt}

\begin{IEEEbiography}[{\includegraphics[width=1in,height=1.25in,clip,keepaspectratio]{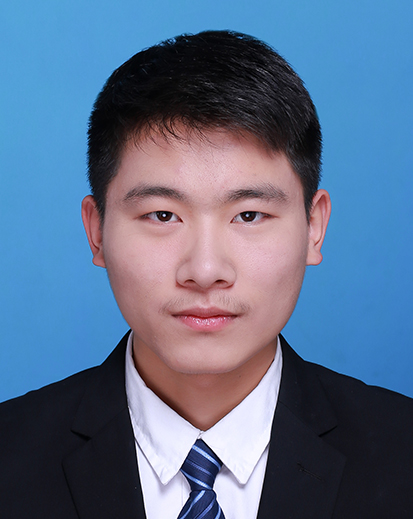}}]{Chuang Wang} received the B.S. degree in School of Mechanical and Power Engineering from Zhengzhou University, China, in 2017. He is currently pursuing the Ph.D. degree with the Shien-Ming Wu School of Intelligent Engineering, South China University of Technology, China. His research interests include robotic manipulation, compliance control, deep reinforcement learning, and assembly robot.
\end{IEEEbiography}

\begin{IEEEbiography}[{\includegraphics[width=1in,height=1.25in,clip,keepaspectratio]{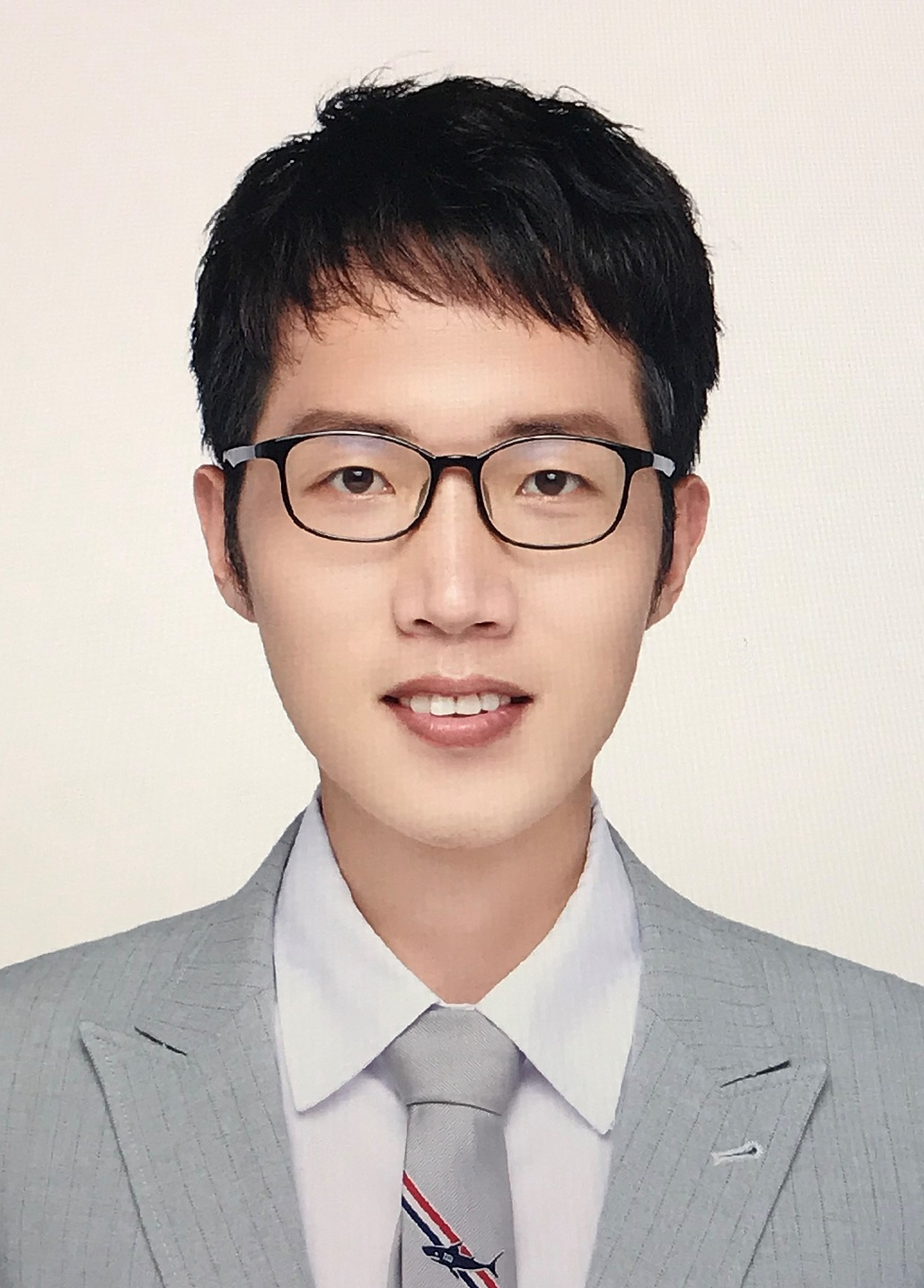}}]{Lie Yang} received the Ph.D. degree from the Shien-Ming Wu School of Intelligent Engineering, South China University of Technology, Guangzhou, China, in 2021. He was a lecturer in the School of Computer Science and Technology, Hainan University, in 2022. He is currently a Research Fellow with the Department of Mechanical and Aerospace Engineering, Nanyang Technological University, Singapore. His research interests mainly focus on deep learning, computer vision, pattern recognition, driver state monitoring, and brain-computer interface.
\end{IEEEbiography}

\begin{IEEEbiography}[{\includegraphics[width=1in,height=1.25in,clip,keepaspectratio]{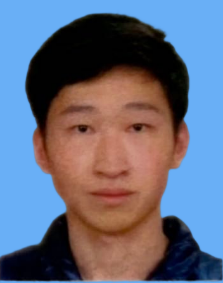}}]{Ze Lin} is currently pursuing the B.S. degree in Intelligent Manufacturing Engineering with the Shien-Ming Wu School of Intelligent Engineering, South China University of Technology, China. His current research interests include robotic manipulation, machine vision and assembly robot.
\end{IEEEbiography}

\begin{IEEEbiography}[{\includegraphics[width=1in,height=1.25in,clip,keepaspectratio]{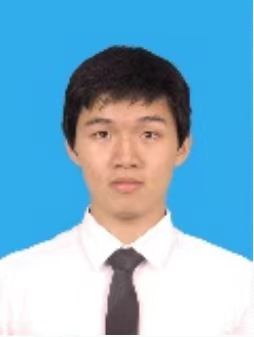}}]{Yizhi Liao} received a M.S. degree in Advanced Computer Science at the University of Sheffield, United Kingdom. Currently, he is pursuing a second M.S. degree in Information Technology at the University of Melbourne, Australia. His current research interests include robotic manipulation and computer vision.
\end{IEEEbiography}

\begin{IEEEbiography}[{\includegraphics[width=1in,height=1.25in,clip,keepaspectratio]{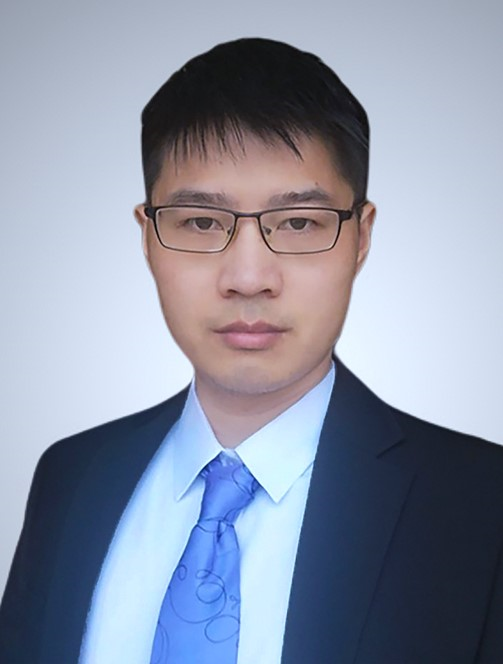}}]{Gang Chen} received the bachelor's and master's degrees in mechanical engineering from Shanghai Jiao Tong University, Shanghai, China, in 2012 and 2015, respectively and his Ph.D. degree in mechanical and aerospace engineering from the University of California, Davis, Davis, CA, in 2020. He was a research fellow with the School of Electrical and Electronic Engineering, Nanyang Technological University, Singapore from 2020 to 2021.  He is currently an associate professor at the Shien-Ming Wu School of Intelligent Engineering, South China University of Technology, China. His research interests include machine learning, formal methods, control, signal processing and fault diagnosis.
\end{IEEEbiography}

\begin{IEEEbiography}[{\includegraphics[width=1in,height=1.25in,clip,keepaspectratio]{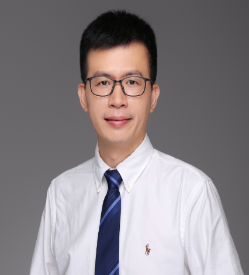}}]{Longhan Xie} received a B.S. degree and a M.S. degree in mechanical engineering in 2002 and 2005, respectively, from Zhejiang University. He received a Ph.D. degree in mechanical and automation engineering in 2010 from the Chinese University of Hong Kong. From 2010 to 2016, he was an Assistant Professor and Associate Professor in the School of Mechanical and Automotive Engineering at the South China University of Technology. Since 2017, he has been a professor in Shien-Ming Wu School of Intelligent Engineering at the same university. His research interests include biomedical engineering and robotics. He is a member of ASME and IEEE.
\end{IEEEbiography}

\vfill

\end{document}